\documentclass[runningheads]{llncs}

\usepackage{eccv}

\usepackage{eccvabbrv}

\usepackage{graphicx}
\usepackage{booktabs}
\usepackage{pifont}
\usepackage{wrapfig}
\usepackage{multirow}
\usepackage{multicol}
\usepackage{kotex}

\usepackage{xspace}
\usepackage{arydshln}
\usepackage{amsmath}
\usepackage{bbm} 
\usepackage{tablefootnote}

\DeclareMathOperator*{\argmin}{arg\,min}

\usepackage[accsupp]{axessibility}  

\newcommand{\cmark}{\textcolor{green}{\ding{51}}}%
\newcommand{\xmark}{\textcolor{red}{\ding{55}}}%

\definecolor{RED}{rgb}{0.86,0.13,0.13}
\definecolor{GREEN}{rgb}{0.0,0.6,0.0}

\usepackage[pagebackref,breaklinks,colorlinks,citecolor=eccvblue]{hyperref}

\usepackage{orcidlink}
\usepackage{xcolor,colortbl}

\begin{document}

\title{DHR: Dual Features-Driven Hierarchical Rebalancing in Inter- and Intra-Class Regions for Weakly-Supervised Semantic Segmentation}

\titlerunning{DHR}

\def\thefootnote{$*$}\footnotetext{Correspondence to}\def\thefootnote{\arabic{footnote}}

\author{
Sanghyun Jo\inst{1} \and
Fei Pan\inst{2} \and 
In-Jae Yu\inst{3} \and
Kyungsu Kim$*$\inst{4}
}

\authorrunning{S. Jo et al.}

\institute{
OGQ, Seoul, Korea \and
University of Michigan, Ann Arbor, MI, USA \and
Samsung Electronics, Suwon, Korea \and
Massachusetts General Hospital and Harvard Medical School, Boston, MA, USA \\
\email{\{shjo.april, ijyu.phd, kskim.doc\}@gmail.com}, \email{feipan@umich.edu}
}

\maketitle

\begin{abstract}
Weakly-supervised semantic segmentation (WSS) ensures high-quality segmentation with limited data and excels when employed as input seed masks for large-scale vision models such as Segment Anything. However, WSS faces challenges related to minor classes since those are overlooked in images with adjacent multiple classes, a limitation originating from the overfitting of traditional expansion methods like Random Walk. We first address this by employing unsupervised and weakly-supervised feature maps instead of conventional methodologies, allowing for hierarchical mask enhancement. This method distinctly categorizes higher-level classes and subsequently separates their associated lower-level classes, ensuring all classes are correctly restored in the mask without losing minor ones. Our approach, validated through extensive experimentation, significantly improves WSS across five benchmarks (VOC: 79.8\%, COCO: 53.9\%, Context: 49.0\%, ADE: 32.9\%, Stuff: 37.4\%), reducing the gap with fully supervised methods by over 84\% on the VOC validation set. Code is available at \url{https://github.com/shjo-april/DHR}.   
\end{abstract}

\section{Introduction} 
\label{sec:intro}

\begin{figure}
\centering
\includegraphics[width=0.95\linewidth]{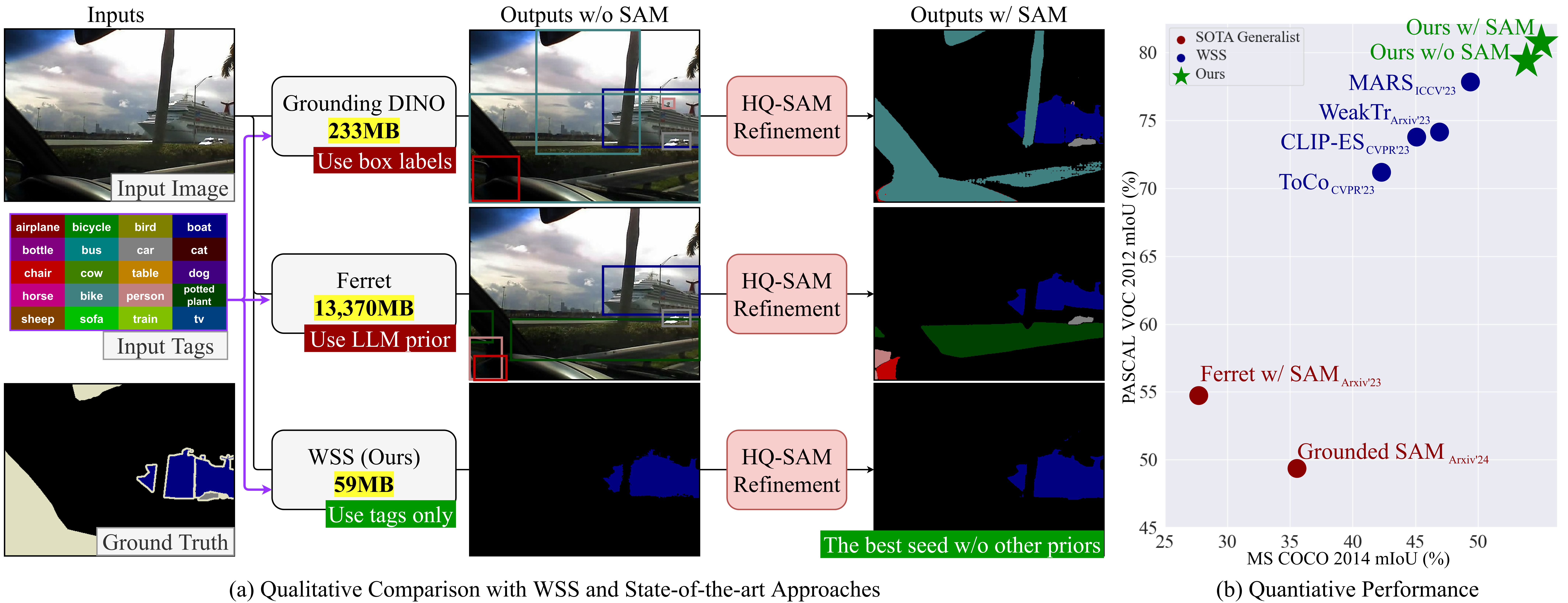}
\vspace{-0.3cm}
\caption{
    \textbf{Importance of WSS.} \textbf{(a):} Our WSS approach (DHR) outperforms large-scale vision models \cite{ren2024grounded, you2023ferret} with only image-level supervision and 25\% of the parameters, bypassing the need for extensive human annotations (image, text, and box pairs). \textbf{(b):} Our DHR significantly exceeds Grounded SAM \cite{ren2024grounded}, Ferret \cite{you2023ferret}, and recent WSS models \cite{ru2023token, Lin_2023_CVPR, zhu2023weaktr, Jo_2023_ICCV} in standard benchmark performances \cite{everingham2010pascal, lin2014microsoft}.
}
\label{fig:importance}
\end{figure}

Semantic segmentation, the process of grouping each pixel in an image into semantic classes, heavily relies on pixel-wise annotations crafted by humans. This labor-intensive requirement has been a significant bottleneck in scaling segmentation models. Recently, large-scale visual models (LSVMs), such as Grounded SAM \cite{ren2024grounded}, have emerged as compelling alternatives in segmentation tasks, employing advanced box, point, or scribble supervision with image-text pairs. By contrast, weakly-supervised semantic segmentation (WSS) utilizes image-level class labels with images to produce segmentation outcomes. As shown in Figure \ref{fig:importance}, recent WSS models \cite{ru2023token,Jo_2023_ICCV} outperform existing LSVMs by at least 7\% accuracy on segmentation datasets, such as VOC 2012 \cite{everingham2010pascal} and COCO 2014 \cite{lin2014microsoft}.

\begin{figure}[!b]
\centering
\includegraphics[width=0.95\linewidth]{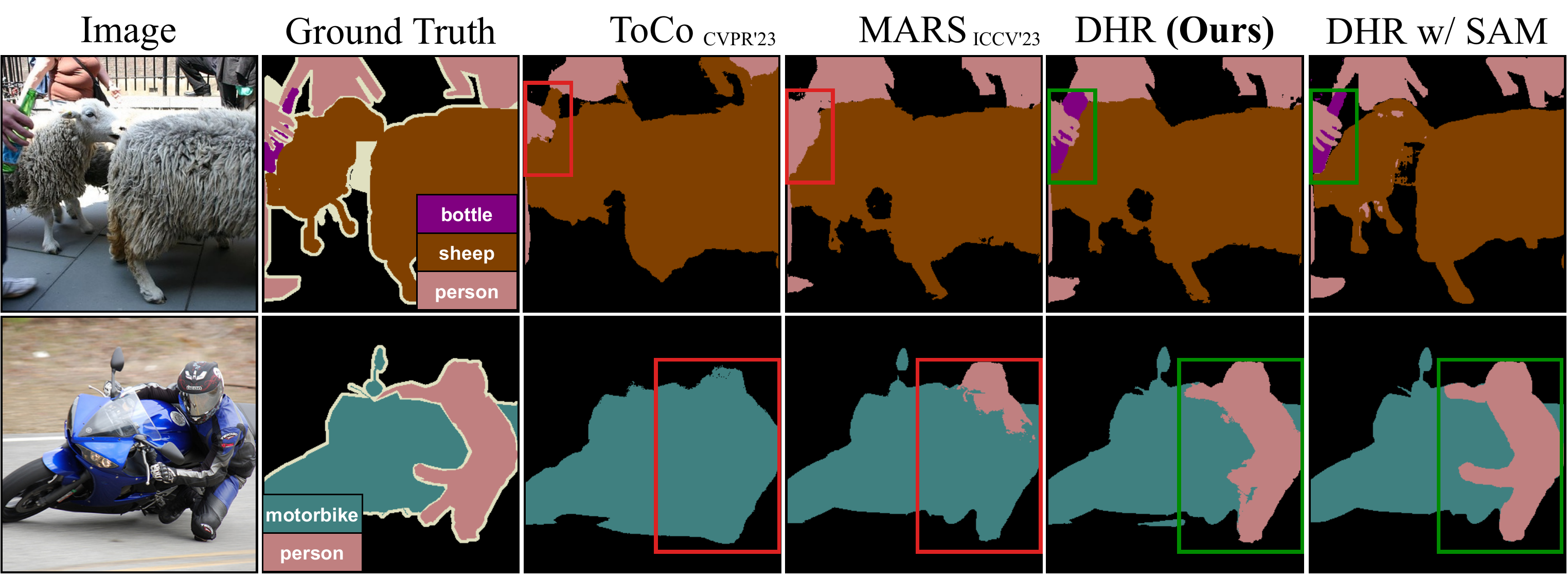}
\vspace{-0.3cm}
\caption{
    \textbf{Vanishing problem of adjacent minor classes in WSS outputs.} Red boxes illustrate the false prediction of minor classes in pseudo labels generated from WSS, \emph{e.g.}, bottle, person, and backpack. Green boxes highlight our DHR outperforming state-of-the-art baselines \cite{ru2023token, Jo_2023_ICCV} in adjacent class regions.
}
\label{fig:problem}
\end{figure}

WSS research efforts \cite{jo2021puzzle, wang2020self, lee2021anti} primarily focus on generating pseudo masks from image-level tags to mimic pixel-level annotations through the propagation of initial class activation maps (CAMs). Our analysis of discrepancies between WSS predictions and ground-truth masks, as shown in Figure \ref{fig:problem}, uncovers a significant challenge: adjacent pixels from distinct classes (\emph{e.g.}, person and motorbike) often merge, leading to the disappearance of spatially minor classes. This problem, a result of overlooking class ratios during the propagation process, is especially pronounced in areas of inter-class regions. Notably, on the VOC dataset \cite{everingham2010pascal}, adjacent regions constitute 35\% of the total area, with 79\% being inter-class regions, while the COCO dataset \cite{lin2014microsoft} has 75\% adjacent regions, 55\% of which are inter-class. This analysis underscores the importance of addressing the vanishing problem in neighboring classes to enhance WSS performance.

\begin{figure}[!t]
\centering
\includegraphics[width=0.90\linewidth]{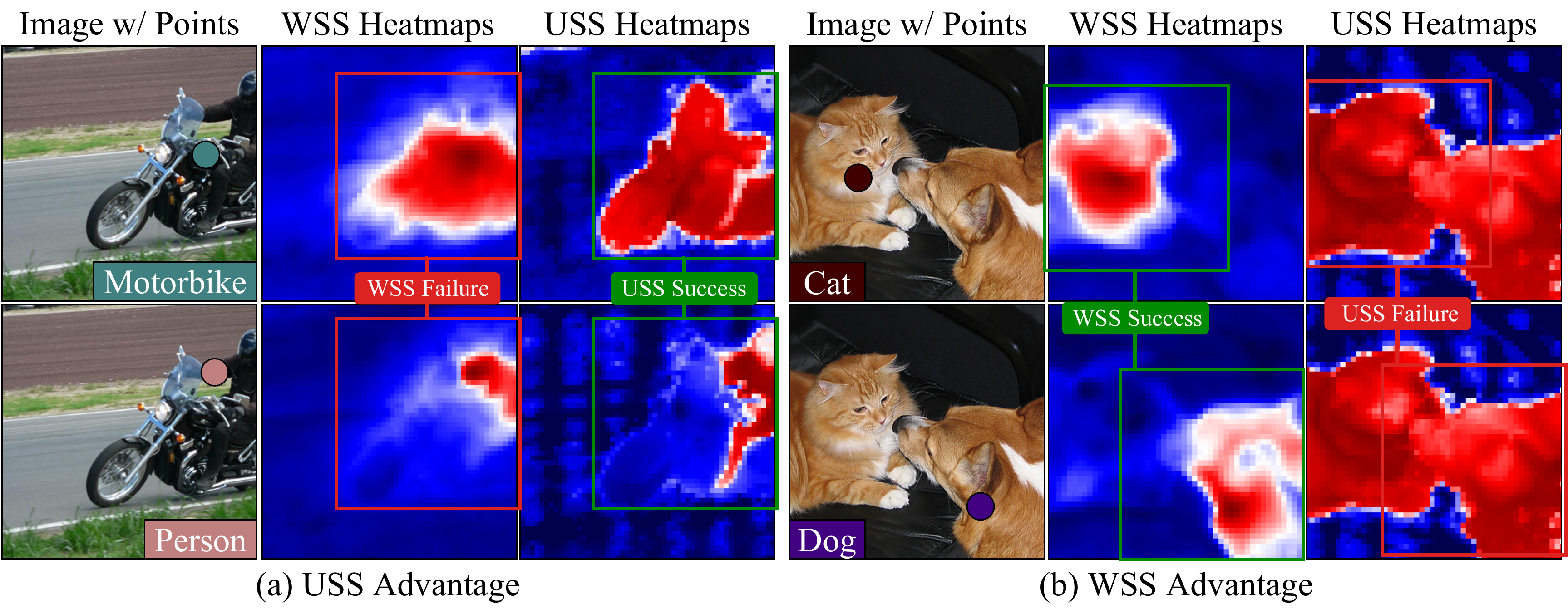}
\vspace{-0.3cm}
\caption{
    \textbf{Visualization of heatmaps with target points.} \textbf{(a):} USS features can precisely separate between inter-class regions (\emph{e.g.}, animals vs. vehicles) unlike WSS. \textbf{(b):} Thanks to training image-level class labels, WSS features can discern specific classes (\emph{e.g.}, dog vs. cat) in the same inter-class region (\emph{e.g.}, animal).
}
\label{fig:motivation}
\end{figure}

We observe that unsupervised semantic segmentation (USS) features, as highlighted in studies \cite{hamilton2022unsupervised, kim2023causal}, are adept at distinguishing between inter-class regions (\emph{e.g.}, animal and furniture). {Figure \ref{fig:motivation} showcases pixel-wise cosine similarity maps (\emph{i.e.}, heatmaps) comparing WSS and USS features with class points. Specifically, the inherent difference between USS and WSS features originates from their distinct learning objectives. In Figure \ref{fig:motivation}(a), USS methods learn visual similarities (\emph{e.g.}, color and shape) between images without tags, enabling them to distinguish inter-class regions with dissimilar appearances (\emph{e.g.}, person vs. motorbike). Conversely, WSS methods learn class-specific differences with human-annotated tags, allowing them to discern visually similar classes (\emph{e.g.}, dog vs. cat) within the same inter-class region (\emph{e.g.}, animals), as shown in Figure \ref{fig:motivation}(b).}

\begin{figure}
\centering
\includegraphics[width=0.90\linewidth]{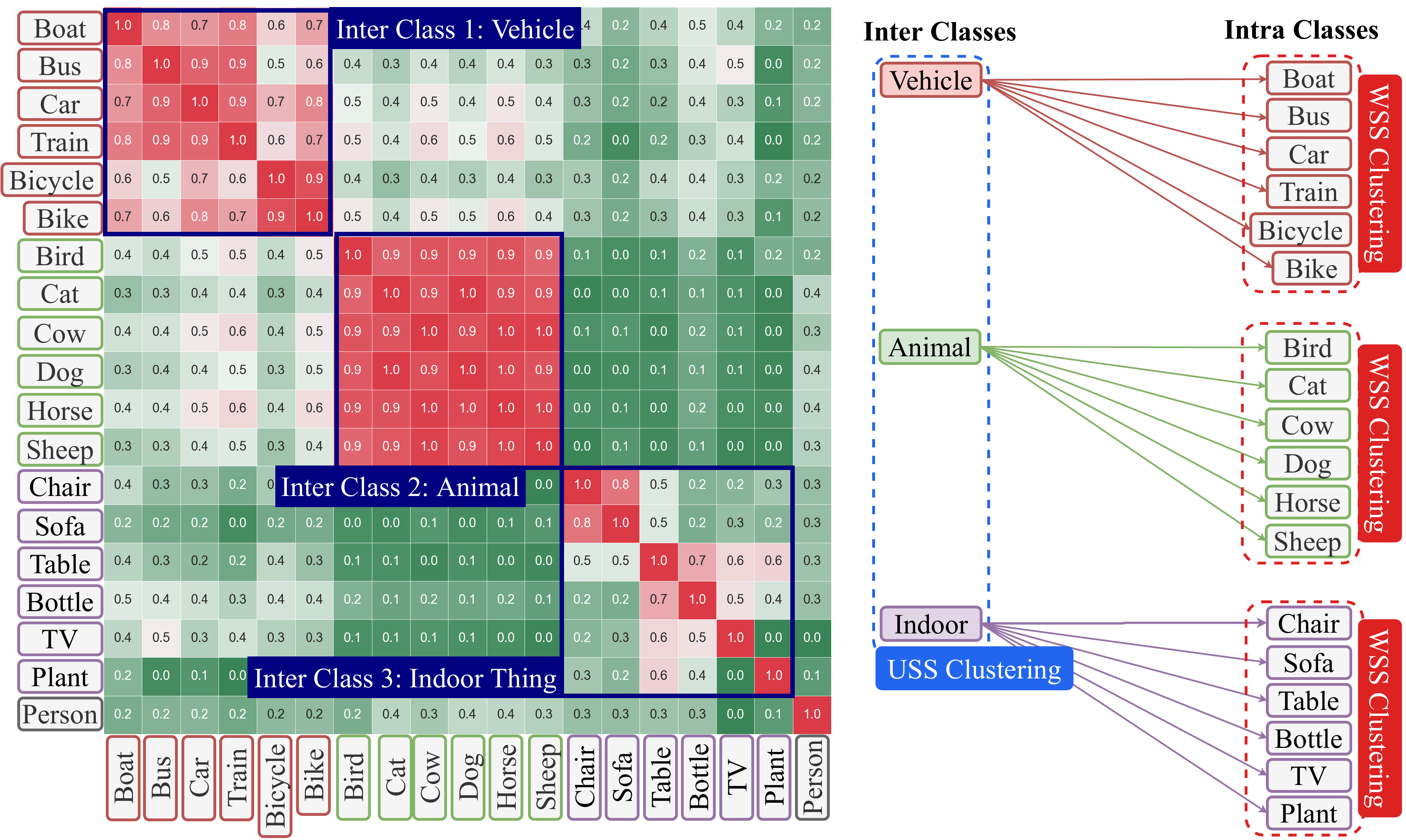}
\vspace{-0.25cm}
\caption{
    \textbf{Conceptual illustration of our hierarchical clustering.} \textit{Left.} USS feature correlation automatically groups inter classes. \textit{Right.} Using USS features, we categorize all inter classes (\emph{e.g.}, vehicle and animal) and then separate these intra classes (\emph{e.g.}, car and bus) per each inter class (\emph{e.g.}, vehicle) with WSS features.
}
\label{fig:idea}
\end{figure}

To harness the strengths of both USS and WSS features, we introduce a pioneering seed propagation method termed \textbf{D}ual Features-Driven \textbf{H}ierarchical \textbf{R}ebalancing (DHR). DHR encompasses three pivotal steps: 1) the seed recovery for disappeared classes in WSS masks, 2) the utilization of USS features for inter-class segregation, and 3) the fine-grained separation between intra-class regions using WSS features. In particular, we automatically group inter- and intra-class regions based on the USS feature correlation matrix to apply subsequent WSS rebalancing in USS-based rebalancing outputs, as visualized in Figure \ref{fig:idea}(a). Consequently, our novel propagation using dual features achieves class separation across all adjacent classes, as conceptually illustrated in Figure \ref{fig:idea}(b). Our key contributions are summarized as follows: 
\begin{itemize}
    \item We identify and tackle the issue of minor classes disappearing in WSS, a problem that previously needed to be addressed by existing techniques.
    \item We introduce DHR, a novel method that 1) enhances class distinction through seed initialization based on optimal transport and 2) leverages USS and WSS features to separate inter- and intra-class regions.
    \item DHR achieves a state-of-the-art mIoU of 79.8\% on the PASCAL VOC 2012 test set, significantly closing the gap with FSS (84\%) and showing versatility across multiple USS and WSS models.
    \item Our method also proves effective as a seeding technique for SAM, outperforming leading approaches like Grounding DINO \cite{liu2023grounding} and establishing the potential of WSS in advanced segmentation tasks.
    \item We first evaluate our DHR and recent WSS methods \cite{ru2023token, Jo_2023_ICCV} across three additional benchmarks \cite{mottaghi2014role, zhou2019semantic, caesar2018coco}. This showcases their adaptability and sets new standards for future explorations in the field.
\end{itemize}


\section{Related Work}
\label{sec:related}

\subsection{Weakly-supervised Semantic Segmentation}
Weakly-supervised semantic segmentation (WSS) aims to minimize the quality gap between pixel-wise annotations and pseudo masks generated with image-level class labels. Unlike previous WSS studies \cite{feng2023weakly, chen2023segment, sun2023alternative, chen2024region, kweon2023weakly, ru2023token, zhu2023weaktr, Lin_2023_CVPR, deng2023qa-clims, yang2024foundation, kim2023semantic, Jo_2023_ICCV}, our research pioneers in addressing the problem of vanishing minor classes by rebalancing class information in pseudo masks in Table \ref{tab:key}. 

\begin{table}
\centering
\caption{\textbf{Conceptual comparison of our method with recent approaches.}}
\vspace{-0.25cm}
\label{tab:key}
\resizebox{\textwidth}{!}{
    \begin{tabular}{@{\extracolsep{1.5pt}}p{0.80\textwidth}|p{0.07\textwidth} p{0.07\textwidth} p{0.10\textwidth} p{0.12\textwidth} p{0.14\textwidth} p{0.07\textwidth} p{0.08\textwidth} p{0.08\textwidth}@{}}
        \toprule
        Properties                           & ACR    & ToCo   & WeakTr & CLIP-ES & QA-CLIMS & FMA    & MARS   & \textbf{Ours} \\
        \midrule
        Solving Vanishing Issue of Adjacent Minor Classes & \xmark & \xmark & \xmark & \xmark  & \xmark  & \xmark & \xmark & \cmark \\
        Utilizing WSS features to improve the quality of WSS masks & \cmark & \cmark & \cmark & \cmark  & \cmark   & \cmark & \xmark & \cmark \\
        Utilizing USS features to enhance the quality of WSS masks & \xmark & \xmark & \xmark & \xmark  & \xmark   & \xmark & \cmark & \cmark \\
        No external datasets and models  & \cmark & \cmark & \cmark & \xmark  & \xmark   & \xmark & \cmark & \cmark \\
        Following model-agnostic manner  & \xmark & \xmark & \xmark & \xmark  & \xmark   & \xmark & \cmark & \cmark \\
        \bottomrule
    \end{tabular}
}
\end{table}

\subsubsection{Evolution of WSS Pipelines} Traditionally, WSS pipelines are composed of three stages: 1) CAM generation for initial seeds, 2) applying propagation methods like Random Walk (RW) \cite{ahn2018learning, ahn2019weakly} to generate pseudo masks from CAMs, and 3) training the final segmentation network (\emph{e.g.}, DeepLabv2 \cite{chen2017deeplab}). Most studies have focused on improving CAM performance using WSS feature correlation \cite{wang2020self,du2022weakly,chen2022self,jo2022recurseed}, patch-based principles \cite{lee2019ficklenet,zhang2021complementary,jo2021puzzle,jiang2022l2g}, and enhancing cross-image information \cite{sun2020mining,wu2021embedded,zhou2022regional,xie2022c2am}. Recent studies \cite{ru2022learning,xu2022multi} employ self-attention mechanisms from transformer architectures rather than the conventional RW for propagating initial CAMs. However, the absence of unsupervised features in these models leads to the disappearance of minor classes during propagation.

\subsubsection{Exploiting Unsupervised Features} A few methods \cite{kim2023semantic, Jo_2023_ICCV} utilize unsupervised features, primarily DINO \cite{caron2021emerging}, for improving WSS performance. MARS \cite{Jo_2023_ICCV} leverages advanced USS features from STEGO \cite{hamilton2022unsupervised} for binary separation of foreground and background classes to remove biased objects in pseudo labels. However, relying solely on unsupervised features without WSS features fails to distinguish between intra-class regions. Our approach is the first solution to mitigate the vanishing problem of adjacent minor classes in WSS outputs by combining WSS and USS features hierarchically, using each of the strengths to separate intra- and inter-class regions simultaneously for the first time.

\subsubsection{Integration with External Models} Current state-of-the-art approaches \cite{Lin_2023_CVPR, deng2023qa-clims, yang2024foundation, chen2024region, chen2023segment, sun2023alternative} rely on CLIP \cite{radford2021learning}, large-language models \cite{you2023ferret}, Grounding DINO \cite{liu2023grounding}, or SAM \cite{kirillov2023segment} to fine-tune WSS models. However, the dependency on pre-trained knowledge limits applicability to novel tasks, such as in medical fields, constraining WSS's scalability. Nevertheless, our approach using both unsupervised and weakly-supervised features outperforms recent WSS methods depending on advanced supervision and datasets, demonstrating the potential of integrating WSS with USS in overcoming limitations presented by external models.

\subsection{Unsupervised Semantic Segmentation}
Unsupervised semantic segmentation (USS) is dedicated to developing semantically rich features across a collection of images without relying on any annotations. Depending on whether self-supervised vision transformers (\emph{e.g.}, \cite{caron2021emerging, oquab2023dinov2}) are used, they can be classified into two types. First, without initializing pre-trained vision transformers, conventional approaches \cite{ji2019invariant, ouali2020autoregressive, cho2021picie} enhance the mutual information across differing perspectives of the same image. Meanwhile, Leopart \cite{ziegler2022self}, STEGO \cite{hamilton2022unsupervised}, HP \cite{seong2023leveraging}, and CAUSE \cite{kim2023causal} employ self-supervised vision transformers for initializing spatially coherent representations of images and train a simple feed-forward network to enhance pixel-level representation. Our technique is designed to be compatible with any USS approach by leveraging pixel-level embedding vectors. Thus, our method operates independently from existing USS frameworks. Based on recent USS models \cite{hamilton2022unsupervised, kim2023causal}, we demonstrate our method's flexibility to consistently improve the performance of existing WSS models in Section \ref{sec:exp}.

\subsection{Open-vocabulary Detection and Segmentation}
Contrastive Language-Image Pre-training (CLIP) \cite{radford2021learning}, trained on 400 million image-text pairs, set a foundation for generating segmentation from free-form text prompts. Despite advancements with models like MaskCLIP \cite{zhou2022extract} and TCL \cite{cha2023learning}, they fall short against specialized segmentation models, such as DeepLabv3+ \cite{zhou2022extract} and Mask2Former \cite{cheng2022masked}, which utilize precise mask annotations for training. Recently, Grounding DINO \cite{liu2023grounding} and Ferret \cite{you2023ferret} have pushed the envelope by integrating box annotations and leveraging language models to enhance detection performance. The introduction of SAM \cite{kirillov2023segment, ke2024segment}, a zero-shot segmentation model using conditional prompts like points, marks a significant stride in employing open-vocabulary tasks, with Grounded SAM \cite{ren2024grounded} leading the charge by merging two capabilities of Grounding DINO \cite{liu2023grounding} and SAM \cite{kirillov2023segment}. Still, these models face challenges in generating accurate seeds from multi-tag inputs due to their reliance on limited tag ranges in training captions. Our analysis reveals that advanced WSS models, including our DHR approach, outperform these open-vocabulary models in multi-class predictions (see Figure \ref{fig:importance} and Table \ref{tab:performance}). Incorporating DHR with SAM for the final refinement also boosts segmentation accuracy, showcasing WSS's indispensable role in navigating multi-tag segmentation scenarios.

\section{Method}
\label{sec:method}

\begin{figure}
\centering
\includegraphics[width=1.0\linewidth]{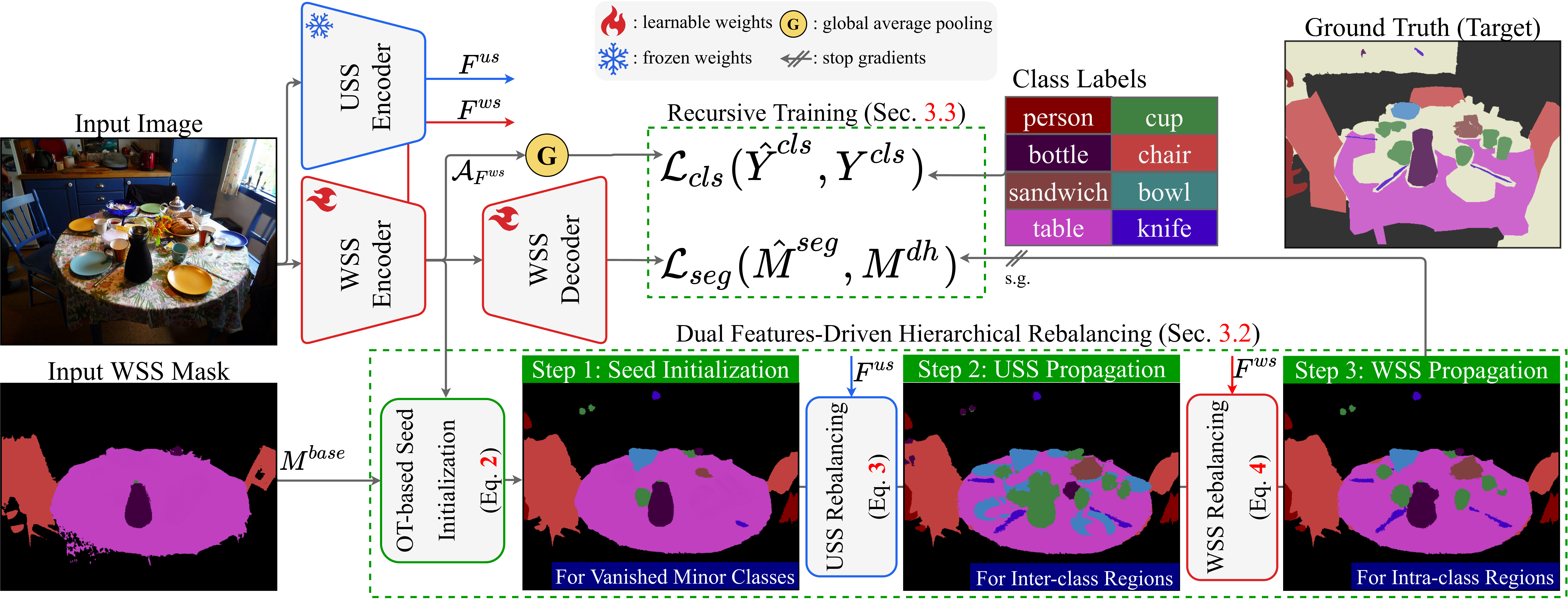}
\vspace{-0.6cm}
\caption{\textbf{Overview of DHR}. Our framework unfolds in three steps, recovering vanished classes by replacing pixels of input WSS mask with OT-based CAMs for seed initialization. We then employ a hierarchical approach to propagate restored seeds, using unsupervised feature maps for the inter-class segregation (\emph{e.g.}, kitchenware) and weakly-supervised features for the intra-class differentiation (\emph{e.g.}, bottle and cup). Finally, our balanced masks are used to train the segmentation model recursively.
}
\label{fig:overview}
\end{figure}

In this section, we introduce our method, DHR (\textbf{D}ual Features-Driven \textbf{H}ierarchical \textbf{R}ebalancing), which aims to address the challenges of vanishing classes in existing WSS methods. We outline an overview of DHR in Figure \ref{fig:overview} for a comprehensive understanding of our framework. Section \ref{sec:background} delves into the background of conventional seed propagation techniques and our setup, highlighting the limitations that our method seeks to overcome. The core of our contribution (DHR) is presented in Section \ref{sec:dhr}, where we detail how we tackle the disappearance of inter- and inter-class regions through a novel rebalancing strategy. We round off with Section \ref{sec:recursive}, discussing DHR's training objectives.

\subsection{Background: Conventional Seed Propagation} \label{sec:background}
To train segmentation models, previous WSS studies \cite{ahn2018learning, jo2022recurseed} generate pseudo masks by expanding initial seeds obtained from class activation maps (CAMs). These CAMs are extracted from weakly-supervised feature maps \(F^{ws} = E_{\theta^{ws}}(I)\), produced by an image encoder within the WSS network, using the target image $I$ as input. Mathematically, the process of propagating CAMs is described as 
\begin{align}\label{conv_mask}
M^{base} = \mathcal{R}_C(\mathcal{P}^{base}(\mathcal{A}_{F^{ws}})),
\end{align}
where \(\mathcal{A}_{F^{ws}}\) represents the CAMs, \(\mathcal{P}^{base}(\cdot)\) indicates conventional seed propagation techniques \cite{huang2018weakly, ahn2019weakly, lee2021reducing} such as Random Walk \cite{ahn2018learning}. The process further incorporates $\mathcal{R}_C(\cdot)$, boundary correction tools such as CRF \cite{krahenbuhl2011efficient} and PAMR \cite{araslanov2020single}, to produce a segmentation output with $C$ class channels, resulting in the final WSS mask $M^{base} \in \mathbb{R}^{C \times H \times W}$. Although this propagation approach covers a sufficient foreground region compared to CAMs, it leads to the disappearance of adjacent minor classes in Figure \ref{fig:problem}.

\subsection{Dual Features-Driven Hierarchical Rebalancing (DHR)}
\label{sec:dhr}

For our DHR to easily integrate with other WSS methods, we aim to refine pre-propagated WSS masks \(M^{base}\) during the segmentation learning phase rather than replacing existing propagation tools. Our DHR approach comprises three steps; this presents a new propagation mechanism \(\mathcal{P}^{ours}(\cdot)\) with reconstructing vanished-class regions.

\subsubsection{Step 1: Optimal Transport-based Seed Initialization}
\label{sec:otseed}
\begin{figure}[h]
\centering
\includegraphics[width=1.0\linewidth]{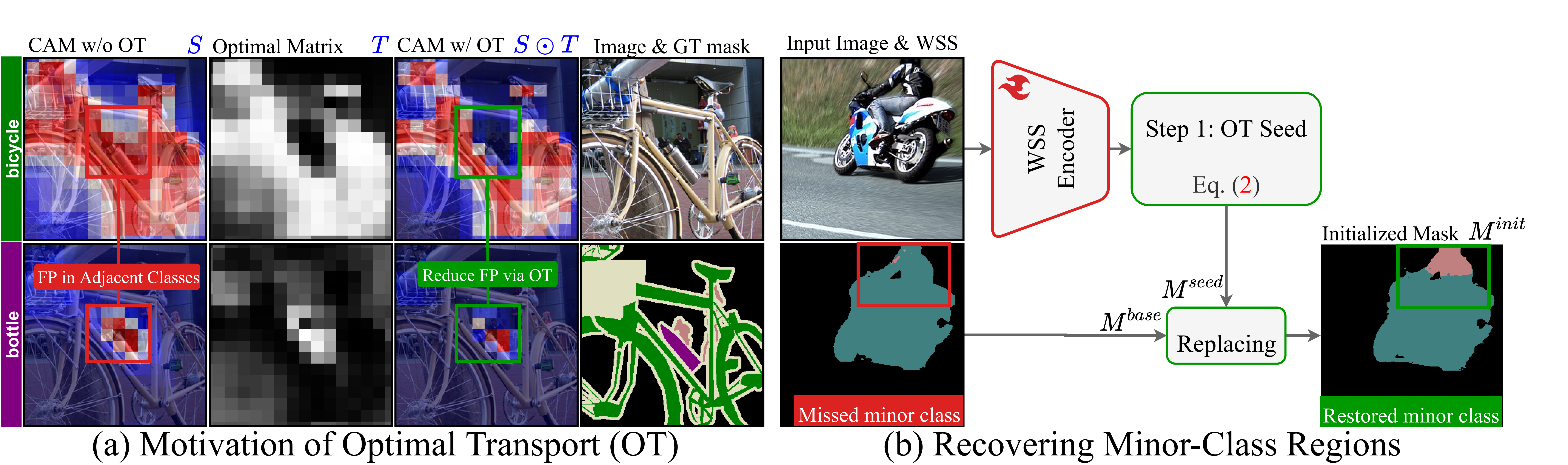}
\vspace{-0.7cm}
\caption{
\textbf{Illustration of recovering minor-class regions from CAMs with OT.} \textbf{(a):} OT reduces false positives in adjacent class regions (\emph{e.g.}, bottle vs. bicycle). \textbf{(b):} The input WSS mask often misses most minor-class areas (\emph{i.e.}, person) in the red box. By contrast, refining CAM seeds with OT eliminates overlapping areas, restoring minor-class regions that have previously vanished, as indicated in the green box.
}
\label{fig:restoring}
\end{figure}

To restore minor-class regions that vanish during the traditional propagation process, we revisit CAMs \(\mathcal{A}_{F^{ws}}\) utilized before the propagation \(\mathcal{P}^{base}\). As shown in Figure \ref{fig:restoring}(a), we first observe that Optimal Transport (OT) \cite{rachev1985monge} effectively minimizes the occurrence of false positives (FP) in regions adjacent to the target classes: 
\begin{align} \label{eq:ot}
M^{seed} = \mathcal{R}_C(f_{OT}(\mathcal{A}_{F^{ws}}) \odot \mathcal{A}_{F^{ws}}),
\end{align}
{where $f_{OT}(S) := \underset{{T}}{\argmin} \sum_{i=1}^{HW}\sum_{j=1}^{C} T_{ij} (1 - S_{ij}) - \lambda \mathcal{H}(T)$ represents the optimal transport matrix \textcolor{blue}{\(T\)} for performing OT based on the input heatmaps \textcolor{blue}{\(S\)} (\emph{e.g.}, CAM). Here, \(\lambda\) is a regularization parameter set to 0.1 and \(\mathcal{H}(\cdot)\) denotes the entropy term. Finally, the recovered minor-class regions from \(M^{seed}\) are integrated with \(M^{base}\) to initialize the WSS mask \(M^{init} \in \mathbb{R}^{C \times H \times W}\) in Figure \ref{fig:restoring}(b).}

\vspace{-0.40cm}

\vspace{-0.80cm}

\subsubsection{Step 2: USS Feature-based Rebalancing For Inter-class Regions}
\begin{figure}[!t]
\centering
\includegraphics[width=\linewidth]{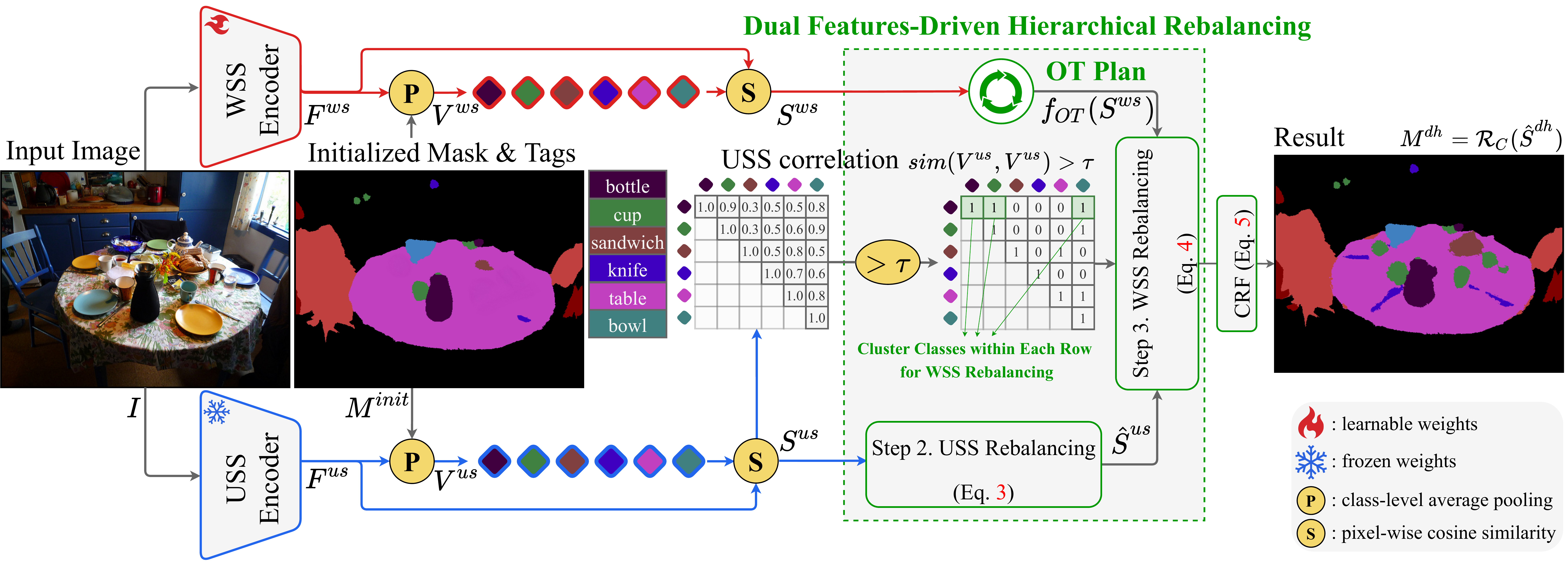}
\vspace{-0.75cm}
\caption{\textbf{Visualization of hierarchical rebalancing based on dual features.} From initialized seeds \(M^{init}\) with restored minor-class regions obtained in the first step of Section \ref{sec:otseed}, USS features are used for inter-class grouping, while WSS features are sequentially applied for intra-class separation within each inter-class region. Here, WSS is conditionally applied to cases with high USS correlation (\emph{i.e.}, classes that USS cannot distinguish), successfully differentiating both inter-class and intra-class pixels.}
\label{fig:rebalancing}
\end{figure}

The bottom part (blue) in Figure \ref{fig:rebalancing} shows the second step of DHR. We obtain unsupervised feature maps \(F^{us}\) from the target image \(I\), defined as \(F^{us} = E_{\theta^{us}}(I)\) with dimensions \(H \times W \times D_{us}\). We apply class-level average pooling (CAP) to \(F^{us}\) using the initial mask \(M^{init}\), producing USS centroids for each class, denoted as \(V^{us} = CAP(F^{us},M^{init})\) with dimensions \(C \times D_{us}\). This leads to the creation of the USS-based mask \(\hat{S}^{us}\):
\begin{align} \label{eq:uss}
\hat{S}^{us} := f_{OT}(S^{us}) \odot S^{us},
\end{align}
where \(S^{us}_{ij} := ReLU(sim(F^{us}_{ij}, V^{us}))\) in \(\mathbb{R}^{C}\) is the result of updating the initial mask $M^{init}$ by clustering around USS centroids \(V^{us}\), aiding in the distinct categorization of inter-class regions (\emph{e.g.}, kitchenware and furniture). $\textit{sim}(\cdot)$ is the cosine similarity. The refined outcome, \(\hat{S}^{us}\), is the result of applying OT-based optimization to this updated mask. We describe CAP in Appendix \ref{sec:cap}. 

\subsubsection{Step 3: WSS Feature-based Rebalancing For Intra-class Regions} 
In the last stage of DHR, we refine the mask \(\hat{S}^{us}\) with weakly-supervised feature maps \(F^{ws} = E_{\theta^{ws}}(I)\), sized \(\mathbb{R}^{H \times W \times D_{ws}}\), for the precise intra-class segregation, as shown in Figure \ref{fig:rebalancing}. Following the second step, we generate WSS centroids \(V^{ws} = CAP(F^{ws},M^{init})\) in \(\mathbb{R}^{C}\) by applying CAP to \(F^{ws}\) with \(M^{init}\). The final refined mask \(\hat{S}^{dh}\) is then formed:
\begin{align} \label{eq:wss}
\hat{S}^{dh} := f_{OT}(S^{ws}) \odot \hat{S}^{us} \odot \mathbbm{1}_{[{sim}(V^{us}, V^{us}) > \tau]}
\end{align}
where \(S^{ws}_{ij} := ReLU({sim}(F^{ws}_{ij}, V^{ws}))\) in \(\mathbb{R}^{C}\) creates a WSS-based clustered mask for the intra-class distinction. Combined with \(\hat{S}^{us}\) through \(f_{OT}(S^{ws}) \odot \hat{S}^{us}\), it could enhance the further differentiation across all class regions. Consequently, the final mask \(\hat{S}^{dh}\) balances both intra- and inter-class distinctions. The pruning operator \(\mathbbm{1}_{[{sim}(V^{us}, V^{us}) > \tau]}\) applies WSS rebalancing for high-correlated USS centroids, thereby leveraging the strengths of WSS and USS features. Lastly, we employ the same boundary correction tool $\mathcal{R}_C(\cdot)$ in \eqref{conv_mask} used in existing WSS studies to generate the final segmentation mask $M^{dh}$: 
\begin{align}\label{our_mask}
M^{dh}:=\mathcal{R}_C(\hat{S}^{dh})  := \mathcal{R}_C(\mathcal{P}^{ours}(\mathcal{A}_{F^{ws}}))  
\end{align}
Contrary to former approaches like \(\mathcal{R}_C(\mathcal{P}^{base}(\mathcal{A}_{F^{ws}}))\) in \eqref{conv_mask}) which apply flawed propagation \(\mathcal{P}^{base}\), our DHR initiates a new propagation, \(\mathcal{P}^{ours}\), incorporating dual feature maps from USS (Step 2) and WSS (Step 3) to discern inter- and intra-class regions for all class rebalancing. We add the specific example related to the pruning operator \(\mathbbm{1}_{[{sim}(V^{us}, V^{us}) > \tau]}\) in Appendix \ref{sec:uss_feature_correlation}.

\subsection{Recursive Learning}
\label{sec:recursive} 
The last section presents our training strategy using refined masks $M^{dh}$. To train both WSS encoder and decoder parameters, we combine two different losses following previous methods \cite{araslanov2020single, ru2022learning, jo2022recurseed}:
\begin{align}
\mathcal{L}_{total} = \mathcal{L}_{cls} (\hat{Y}^{cls}, Y^{cls}) + \mathcal{L}_{seg} (\hat{M}^{seg}, M^{dh})
\end{align}
where $\mathcal{L}_{cls}$ and $\mathcal{L}_{seg}$ denote the multi-label soft margin loss and the per-pixel cross-entropy loss, respectively. We apply global average pooling (GAP) and sigmoid $\sigma$ to predict class labels, denoted as $\hat{Y}^{cls} = \sigma(GAP(\mathcal{A}_{F^{ws}}))$. The final segmentation output $\hat{M}^{seg} = D_{\theta^{ws}}(F^{ws})$ is obtained from the WSS decoder. As a result, our DHR iteratively refines initial WSS masks, utilizing WSS features to separate inter- and intra-class regions. This process has continuously enhanced the quality of WSS outputs through recursive updates by the WSS network.


\section{Experiments}
\label{sec:exp}

\vspace{-0.20cm}

\subsection{Experimental Setup}

\vspace{-0.20cm}

\subsubsection{Datasets}
We conduct all experiments on five segmentation benchmarks, such as PASCAL VOC 2012 (VOC) \cite{everingham2010pascal}, MS COCO 2014 (COCO) \cite{lin2014microsoft}, Pascal Context (Context) \cite{mottaghi2014role}, ADE 2016 (ADE) \cite{zhou2019semantic}, and COCO-Stuff (Stuff) \cite{caesar2018coco}, beyond limited WSS benchmarks. All datasets include image-level class labels and pixel-wise annotations to quantify the performance gap between FSS and WSS. PASCAL VOC 2012 \cite{everingham2010pascal}, MS COCO 2014 \cite{lin2014microsoft}, Pascal Context \cite{mottaghi2014role}, ADE 2016 \cite{zhou2019semantic}, and COCO-Stuff \cite{caesar2018coco} datasets have 21, 81, 59, 150, and 171 classes, respectively. For a fair comparison, we reproduce previous state-of-the-art WSS models \cite{ru2023token, Jo_2023_ICCV} on five segmentation benchmarks.

\vspace{-0.2cm}

\subsubsection{Reproducibility}
For our experiments, all unsupervised semantic segmentation (USS) models \cite{hamilton2022unsupervised, kim2023causal} are trained from the ground up on individual datasets without incorporating additional data. To assess our method's adaptability, we apply our method (DHR) across various weakly-supervised semantic segmentation (WSS) studies \cite{wang2020self, lee2021anti, jo2022recurseed} on the PASCAL VOC 2012 dataset \cite{everingham2010pascal}. We adhere to the original training protocols of both USS and WSS methods to ensure an equitable comparison. Consequently, our approach has the same evaluation runtime as competing methods. For preliminary results, we generate initial WSS masks from  RSEPM \cite{jo2022recurseed} with MARS \cite{Jo_2023_ICCV} and set $\tau$ with 0.8 in \eqref{eq:wss}. In line with standard evaluation practices for segmentation tasks, we employ multi-scale inference and Conditional Random Fields (CRF) \cite{krahenbuhl2011efficient} for producing segmentation outcomes. All experiments are conducted on a single RTX A100 GPU (80GB) using PyTorch for WSS and USS model implementations.

\vspace{-0.2cm}

\subsubsection{Evaluation Metrics}
Our method is evaluated through the mean Intersection over Union (mIoU) metric, aligning with the standard evaluation criterion utilized in previous WSS research \cite{ahn2019weakly, wang2020self, lee2021anti, jo2022recurseed, Jo_2023_ICCV}. We acquire results for the PASCAL VOC 2012 validation and test datasets directly from the official PASCAL VOC online evaluation platform.

\vspace{-0.25cm}

\subsection{Comparison with State-of-the-art Approaches}
We compare our method with other WSS methods based on CNN and transformer architectures for the quantitative analysis. While recent state-of-the-art methods exploit various supervisions, ranging from LLM \cite{deng2023qa-clims} to SAM \cite{yang2024foundation}, our approach achieves significant results solely by depending on image-level supervision. This outcome underscores the effectiveness of leveraging both USS and WSS features to address vanishing adjacent minor classes. Notably, our method exhibits significant performance gains, surpassing the previous state-of-the-art approach \cite{Jo_2023_ICCV} by a margin of at least 1.9\% for the PASCAL VOC 2012 validation and test sets utilizing the same configuration. Surprisingly, the proposed method consistently demonstrates high performance across five benchmarks, validating the efficacy of WSS techniques compared to open-vocabulary models \cite{ren2024grounded}. Extended results, \emph{e.g.}, per-class comparisons, are provided in Appendix \ref{sec:additional_quantitative_results}.

\begin{table}[t]
    \centering
    \caption{Performance comparison of WSS methods across five benchmarks.}
  \vspace{-0.3cm}
\resizebox{\textwidth}{!}{
    
    \begin{tabular}{p{0.60\textwidth}cccccccc}
    \toprule
    \multicolumn{1}{c}{} &  &  & \multicolumn{2}{c}{VOC} & COCO & Context & ADE & Stuff \\  
    \multicolumn{1}{l}{\multirow{-2}{*}{Method}} & \multirow{-2}{*}{Backbone} & \multirow{-2}{*}{Supervision} & val & test & val & val & val & val \\
    \midrule
    \multicolumn{9}{l}{\textbf{Weakly-supervised Segmentation Models:}} \\
    RIB {\tiny NeurIPS'21} \cite{lee2021reducing} & ResNet-101 & $\mathcal{I}$ & 68.3 & 68.6 & 43.8 & - & - & - \\
    URN {\tiny AAAI'22} \cite{li2022uncertainty} & ResNet-101 & $\mathcal{I}$ & 69.5 & 69.7 & 40.7 & - & - \\
    W-OoD {\tiny CVPR'22} \cite{lee2022weakly} & ResNet-101 & $\mathcal{I}$+$\mathcal{D}$ & 69.8 & 69.9 & -  & - & - & - \\
    Feng \emph{et al.} {\tiny Pattern Recognition'23} \cite{feng2023weakly} & ResNet-101 & $\mathcal{I}$ & 70.5 & 71.8 & - & - & - \\
    SANCE {\tiny CVPR'22} \cite{li2022towards} & ResNet-101 & $\mathcal{I}$ & 70.9 & 72.2 & 44.7 & - & - & - \\
    SEPL {\tiny arXiv'23} \cite{chen2023segment} & ResNet-101 & $\mathcal{I}$+$\mathcal{S}$+$\mathcal{C}$ & 71.1 & - & - & - & - \\
    MCTformer {\tiny CVPR'22} \cite{xu2022multi} & Wide-ResNet-38 & $\mathcal{I}$ & 71.9 & 71.6 & 42.0 & - & - & - \\
    L2G {\tiny CVPR'22} \cite{jiang2022l2g} & ResNet-101 & $\mathcal{I}$+$\mathcal{A}$ & 72.1 & 71.7 & 44.2  & - & - & - \\
    RCA {\tiny CVPR'22} \cite{zhou2022regional} & ResNet-101 & $\mathcal{I}$+$\mathcal{A}$ & 72.2 & 72.8 & 36.8*  & - & - & - \\
    PPC {\tiny CVPR'22} \cite{du2022weakly} & ResNet-101 & $\mathcal{I}$+$\mathcal{A}$ & 72.6 & 73.6 & - & - & - & - \\
    SAS {\tiny AAAI'23} \cite{kim2023semantic} & ResNet-101 & $\mathcal{I}$ & 69.5 & 70.1 & 44.8 & - & - & - \\
    ToCo {\tiny CVPR'23} \cite{ru2023token} & ViT-B &  $\mathcal{I}$ & 71.1 & 72.2 & 42.3 & 25.0* & 10.5* & 14.2* \\
    Jiang et al {\tiny arXiv'23} \cite{jiang2023segment} & ResNet-101 & $\mathcal{I}$+$\mathcal{S}$ & 71.1 & 72.2 & - & - & - & - \\
    ACR {\tiny CVPR'23} \cite{kweon2023weakly} & Wide-ResNet-38 &  $\mathcal{I}$ & 71.9 & 71.9 & 45.3 & - & - & - \\
    BECO {\tiny CVPR'23} \cite{Rong_2023_CVPR} & ResNet-101 &  $\mathcal{I}$ & 72.1 & 71.8 & - & - & - & - \\
    MMSCT {\tiny CVPR'23} \cite{xu2023learning} & Wide-ResNet-38 & $\mathcal{I}$+$\mathcal{C}$ & 72.2 & 72.2 & 45.9 & - & - & - \\
    QA-CLIMS {\tiny MM'23} \cite{deng2023qa-clims} & ResNet-101 & $\mathcal{I}$+$\mathcal{L}$ & 72.4 & 72.3 & 43.2 & - & - & - \\
    OCR {\tiny CVPR'23} \cite{cheng2023out} & Wide-ResNet-38 & $\mathcal{I}$ & 72.7 & 72.0 & 42.5 & - & - & - \\
    CLIP-ES {\tiny CVPR'23} \cite{Lin_2023_CVPR} & ResNet-101 & $\mathcal{I}$+$\mathcal{C}$ & 73.8 & 73.9 & 45.4 & - & - & - \\
    BECO {\tiny CVPR'23} \cite{Rong_2023_CVPR} & MiT-B2 &  $\mathcal{I}$ & 73.7 & 73.5 & 45.1 & - & - & - \\
    WeakTr {\tiny arXiv'23} \cite{zhu2023weaktr} & DeiT-S &  $\mathcal{I}$ & 74.0 & 74.1 & 46.9 & - & - & - \\
    ROSE {\tiny Information Fusion'24} \cite{chen2024region} & ResNet-101 & $\mathcal{I}$+$\mathcal{S}$+$\mathcal{C}$ & 75.4 & 76.6 & 48.3 & - & - & - \\
    Sun \emph{et al.} {\tiny arXiv'23} \cite{sun2023alternative} & ResNet-101 & $\mathcal{I}$+$\mathcal{S}$+$\mathcal{D}$ & 77.2 & 77.1 & 55.6 & - & - & - \\
    FMA-WSSS {\tiny WACV'24} \cite{yang2024foundation} & ResNet-101 & $\mathcal{I}$+$\mathcal{S}$+$\mathcal{C}$ & 77.3 & 76.7 & 48.6 & - & - & - \\
    MARS {\tiny ICCV'23} \cite{Jo_2023_ICCV} & ResNet-101 &  $\mathcal{I}$ & 77.7 & 77.2 & 49.4 & 39.8* & 22.0* & 35.7* \\
    \rowcolor[HTML]{F4CCCC} 
    \multicolumn{1}{l}{\cellcolor[HTML]{F4CCCC}\textbf{DHR (Ours, DeepLabv3+)}} & ResNet-101 &  $\mathcal{I}$ & \textbf{79.6}\tablefootnote{\href{http://host.robots.ox.ac.uk:8080/anonymous/A4RUI9.html}{http://host.robots.ox.ac.uk:8080/anonymous/A4RUI9.html}}  & \textbf{79.8} \tablefootnote{\href{http://host.robots.ox.ac.uk:8080/anonymous/HICQUU.html}{http://host.robots.ox.ac.uk:8080/anonymous/HICQUU.html}} & \textbf{53.9} & \textbf{49.0} & \textbf{32.9} & \textbf{37.4} \\
   \rowcolor[HTML]{F4CCCC} 
   \multicolumn{1}{l}{\cellcolor[HTML]{F4CCCC}\textbf{DHR (Ours, Mask2Former)}} & Swin-L &  $\mathcal{I}$ & \textbf{82.3} & \textbf{82.3} & \textbf{56.8} & \textbf{53.6} & \textbf{36.9} & \textbf{41.1} \\
   \textcolor{gray}{ Upper Bound (DeepLabv3+) {\tiny CVPR'18} \cite{chen2018encoder}} &\textcolor{gray}{ ResNet-101} &\textcolor{gray}{  $\mathcal{M}$} &\textcolor{gray}{ 80.6*} &\textcolor{gray}{ 81.0*} &\textcolor{gray}{ 61.8*} &\textcolor{gray}{ 54.6*} &\textcolor{gray}{ 45.3*} &\textcolor{gray}{ 44.2*} \\
   \textcolor{gray}{Upper Bound (Mask2Former) {\tiny CVPR'22} \cite{cheng2022masked}} &\textcolor{gray}{ Swin-L} & \textcolor{gray}{ $\mathcal{M}$} & \textcolor{gray}{86.0} & \textcolor{gray}{86.1} & \textcolor{gray}{66.7} & \textcolor{gray}{64.3*} & \textcolor{gray}{55.5*} & \textcolor{gray}{50.6*} \\
    \midrule
    \multicolumn{9}{l}{\textbf{Open-vocabulary Segmentation Models:}} \\
    MaskCLIP {\tiny ECCV'22} \cite{zhou2022extract} & ViT-B & $\mathcal{C}$+$\mathcal{T}$  & 29.3 & - & 15.5 & 21.1 & 10.8 & 14.7 \\
    TCL {\tiny CVPR'23} \cite{cha2023learning} & ViT-B & $\mathcal{C}$+$\mathcal{T}$  & 55.0 & - & 33.2 & 33.8 & 15.6 & 22.4 \\
    Ferret {\tiny arXiv'23} \cite{you2023ferret} w/ SAM \cite{ke2024segment} & ViT-H & $\mathcal{B}$+$\mathcal{T}$+$\mathcal{S}$+$\mathcal{L}$ & 54.7 & - & 27.7 & 22.4 & 7.6 & 12.6 \\
    Grounded SAM {\tiny arXiv'24} \cite{ren2024grounded} & Swin-B &$\mathcal{B}$+$\mathcal{T}$+$\mathcal{S}$ & 46.3 & - & 35.7 & 28.1 & 4.8 & 18.8 \\
    \bottomrule
*: we reproduce all results for a fair comparison & & & & & & & & \\ 
$\mathcal{I}$: image-level supervision & \multicolumn{3}{l}{$\mathcal{T}$: text supervision (image-text pairs)} & & & \multicolumn{2}{l}{$\mathcal{A}$: saliency \cite{hou2017deeply}} & \\
$\mathcal{L}$: language supervision (\emph{e.g.}, LLM \cite{you2023ferret}) & \multicolumn{2}{l}{$\mathcal{M}$: mask supervision}& \multicolumn{2}{l}{$\mathcal{S}$: SAM \cite{ke2024segment}} & & \multicolumn{2}{l}{$\mathcal{C}$: CLIP \cite{radford2021learning}} \\ $\mathcal{B}$: box supervision & \multicolumn{2}{l}{$\mathcal{D}$: using the external dataset \cite{lee2022weakly}} \\
    
\end{tabular}
}
  \label{tab:performance}
\end{table}

\subsection{Discussion}

\subsubsection{Flexibility}
We demonstrate the flexibility of our method by comparing it to various WSS and USS methods on the PASCAL VOC 2012 validation dataset. Table \ref{tab:uss} presents the results of incorporating two USS methods, such as STEGO \cite{hamilton2022unsupervised} and CAUSE \cite{kim2023causal}, into the WSS method (\emph{i.e.}, RSEPM \cite{jo2022recurseed}). Both techniques demonstrate a comparable performance improvement of approximately 5\%. This observation suggests that if USS methods can sufficiently ensure performance in distinguishing inter-class regions rather than detailed class distinctions, ample efficacy exists in separating intra-class regions through the proposed WSS rebalancing. We visualize qualitative improvements in Appendix \ref{sec:qualitative_improvements}.

In Table \ref{tab:wss}, we compare our method to other model-agnostic WSS approaches \cite{lee2022weakly,Jo_2023_ICCV,liu2022adaptive} based on four WSS methods \cite {ahn2019weakly, wang2020self,lee2021anti,jo2022recurseed}. We employ CAUSE \cite{kim2023causal} for our USS method as it shows the best performance in Table \ref{tab:uss}. The results demonstrate that, regardless of the specific WSS method employed, our approach consistently outperformed others while significantly narrowing the performance gap between WSS baselines and its subsequent FSS models.

\begin{table}[!t]
    \centering
  \caption{ 
    Comparison with two USS methods in terms of mIoU (\%) on the PASCAL VOC 2012 validation set.
  }
  \vspace{-0.3cm}
  \begin{scriptsize}
  \begin{tabular}{ >{\centering}p{0.2\textwidth} >{\centering}p{0.3\textwidth} >{\centering}p{0.2\textwidth} | c }
    \toprule
    WSS      & USS & Backbone     & mIoU  \\
    \midrule
    RSEPM \cite{jo2022recurseed} & \xmark & ResNet-101 & 74.4 \\
    + MARS \cite{Jo_2023_ICCV} & STEGO {\tiny ICLR'22} \cite{hamilton2022unsupervised} & ResNet-101 & 77.7 (+3.3) \\
\rowcolor[HTML]{F4CCCC} 
+ DHR (Ours) & STEGO {\tiny ICLR'22} \cite{hamilton2022unsupervised} & ResNet-101 & \textbf{79.2 (+4.8)}\\
\rowcolor[HTML]{F4CCCC} 
+ DHR (Ours) & CAUSE {\tiny Arxiv'23} \cite{kim2023causal} & ResNet-101 & \textbf{79.6 (+5.2)}\\
    \bottomrule
  \end{tabular}
  \label{tab:uss}
  \end{scriptsize}
\end{table}

\begin{table}
  \centering
  \caption{Comparison with four WSS methods on the PASCAL VOC 2012 validation set. $\Delta$ means the percentage improvement in the gap between WSS and FSS.
  }
  \vspace{-0.3cm}
  \begin{scriptsize}

\begin{tabular}{p{0.35\textwidth} >{\centering}p{0.25\textwidth} >{\centering}p{0.25\textwidth} | cc}
    \toprule
    WSS & Backbone & Segmentation & mIoU & $\Delta$ \\
    \midrule
    IRNet \cite{ahn2019weakly} & ResNet-50 & DeepLabv2 & 63.5 & 0\% \\
    + MARS \cite{Jo_2023_ICCV} & ResNet-50 & DeepLabv2 & 69.8 & 46\% \\
    \rowcolor[HTML]{F4CCCC}
    + DHR (Ours) & ResNet-50 & DeepLabv2 & \textbf{72.7} & \textbf{72\%} \\
    \textcolor{gray}{Upper Bound (FSS)} & \textcolor{gray} {ResNet-50} & \textcolor{gray}{DeepLabv2} & \textcolor{gray}{76.3} & \textcolor{gray}{-}\\
    \midrule
    SEAM \cite{wang2020self} & Wide-ResNet-38 & DeepLabv1 & 64.5 & 0\% \\
    
    + ADELE \cite{liu2022adaptive} & Wide-ResNet-38 & DeepLabv1 & 69.3 & 35\% \\
    + MARS\cite{Jo_2023_ICCV} & Wide-ResNet-38 & DeepLabv1 & 70.8 & 46\% \\
    \rowcolor[HTML]{F4CCCC} + DHR (Ours) & Wide-ResNet-38 & DeepLabv1 & \textbf{73.6} & \textbf{67\%} \\
   \textcolor{gray}{Upper Bound (FSS)} &\textcolor{gray}{ Wide-ResNet-38} &\textcolor{gray}{ DeepLabv1} &\textcolor{gray}{ 78.1} &\textcolor{gray}{ - } \\
    \midrule
    AdvCAM \cite{lee2021anti} & ResNet-101 & DeepLabv2 & 68.1 & 0\% \\
    + W-OoD \cite{lee2022weakly} & ResNet-101 & DeepLabv2 & 69.8 & 17\% \\
    + MARS \cite{Jo_2023_ICCV} & ResNet-101 & DeepLabv2 & 70.3 & 22\% \\
    \rowcolor[HTML]{F4CCCC} 
    + DHR (Ours) & ResNet-101 & DeepLabv2 & \textbf{73.6} & \textbf{56\%} \\
   \textcolor{gray}{Upper Bound (FSS)} &\textcolor{gray}{ ResNet-101} &\textcolor{gray}{ DeepLabv2} &\textcolor{gray}{ 78.0} &\textcolor{gray}{ -} \\
    \midrule
    RSEPM \cite{jo2022recurseed} & ResNet-101 & DeepLabv3+ & 74.4 & 0\% \\
    + MARS \cite{Jo_2023_ICCV} & ResNet-101 & DeepLabv3+ & 77.7 & 53\% \\
    \rowcolor[HTML]{F4CCCC} 
    + DHR (Ours) & ResNet-101 & DeepLabv3+ & \textbf{79.6} & \textbf{84\%} \\
   \textcolor{gray}{Upper Bound (FSS)} &\textcolor{gray}{ ResNet-101} &\textcolor{gray}{ DeepLabv3+} &\textcolor{gray}{ 80.6} &\textcolor{gray}{ -} \\
    \bottomrule
\end{tabular}

  \label{tab:wss}
  \end{scriptsize}
\end{table}
\vspace{-0.5cm}

\subsubsection{Effect of DHR}
To verify the effectiveness of the main components of the proposed method, we evaluate the quality of pseudo masks when applying OT-based seed initialization and dual features-based hierarchical rebalancing (DHR) (see Section \ref{sec:dhr}) to different segmentation datasets in Table \ref{tab:component}. First, the application of OT-based seed initialization alone enhances the performance by recovering vanished classes in pseudo masks in the second row of Table \ref{tab:component}. Second, the addition of DHR substantiated a notably more efficacious impact. Significantly, in the Context and ADE datasets, composed solely of adjacent classes, our approach demonstrated a minimum performance improvement of over 13.6\%.

Table \ref{tab:ablation} presents a more detailed ablation study of the key components of the proposed method on the COCO training dataset, as the COCO dataset has the largest number of classes (\emph{i.e.}, 81) among the official WSS benchmarks \cite{everingham2010pascal, lin2014microsoft}. We describe a detailed analysis of our components for other training datasets in \ref{sec:consistent_improvements}. Applying OT without boundary correction tools like CRF \cite{krahenbuhl2011efficient} results in a performance increase of +1.7\% (the second row), while the introduction of CRF \cite{krahenbuhl2011efficient} for seed initialization contributed an additional +0.8\% (the third row). The joint application of OT and CRF achieved an overall improvement of +2.5\% (the fourth row) in mIoU compared to our baseline (\emph{i.e.}, MARS \cite{Jo_2023_ICCV}). For DHR, employing USS balancing to refine WSS masks with OT-based seed initialization yields a modest improvement of +1.7\% (the fifth row). However, the combined application of USS and WSS demonstrated a synergistic effect, resulting in a substantial improvement of +6.0\% in the last row, emphasizing the efficacy of their complementary integration. Notably, applying WSS balancing only to WSS masks yields a marginal improvement of 0.2\% (the sixth row), underscoring the correctness of the sequential application of USS followed by WSS. Qualitative improvements are elaborated upon in Appendix \ref{sec:qualitative_improvements}.

\begin{table}[t]
  \centering
  \caption{Effect of key components. We evaluate the quality of pseudo masks (mIoU) on all training datasets.}
  \vspace{-0.3cm}
  \begin{scriptsize}
  \resizebox{\textwidth}{!}{

\begin{tabular}{c c | c c c c c}
    \toprule
\textbf{OT-based Seed Init.} & \textbf{DHR} & VOC & COCO & Context & ADE & Stuff \\
\midrule
\xmark & \xmark & 81.8 & 52.6 & 51.3 & 30.2 & 50.0 \\
\cmark & \xmark & 82.6 (+0.8) & 55.1 (+2.5) & 53.6 (+2.3) & 32.4 (+2.2) & 50.6 (+0.6) \\
\rowcolor[HTML]{F4CCCC} 
\cmark & \cmark & \textbf{83.9 (+2.1)} & \textbf{58.6 (+6.0)} & \textbf{64.9 (+13.6)} & \textbf{48.1 (+17.9)} & \textbf{53.7 (+3.7)} \\
    \bottomrule

\end{tabular}
}
  \label{tab:component}
  \end{scriptsize}
\end{table}

\begin{table}[t]
  \centering
  \caption{Detailed analysis of each component on the MS COCO 2014 training dataset.}
  \vspace{-0.3cm}
  \begin{scriptsize}

\begin{tabular}{>{\centering}p{0.20\textwidth} >{\centering}p{0.20\textwidth} | >{\centering}p{0.20\textwidth} >{\centering}p{0.20\textwidth} | p{0.15\textwidth}}
    \toprule
    \multicolumn{2}{c|}{\textbf{OT-based Seed Initialization}} & \multicolumn{2}{c|}{\textbf{DHR}} & \multirow{2}{*}{mIoU} \\
    Optimal Transport & Refinement & USS Balancing & WSS Balancing &  \\
    \midrule
    \xmark & \xmark & \xmark & \xmark & 52.6 \\
    \cmark & \xmark & \xmark & \xmark & 54.3 (+1.7) \\
    \xmark & \cmark  & \xmark & \xmark & 53.4 (+0.8) \\
    \cmark & \cmark & \xmark & \xmark & 55.1 (+2.5) \\
    \midrule
    \cmark & \cmark & \cmark & \xmark & 56.8 (+4.2) \\
    \cmark & \cmark & \xmark & \cmark & 55.3 (+2.7) \\
    \rowcolor[HTML]{F4CCCC} 
    \cmark & \cmark & \cmark & \cmark & \textbf{58.6 (+6.0)} \\
    \bottomrule
\end{tabular}

  \label{tab:ablation}
  \end{scriptsize}
\end{table}

\begin{figure}[!b]
\centering
\includegraphics[width=0.8\linewidth]{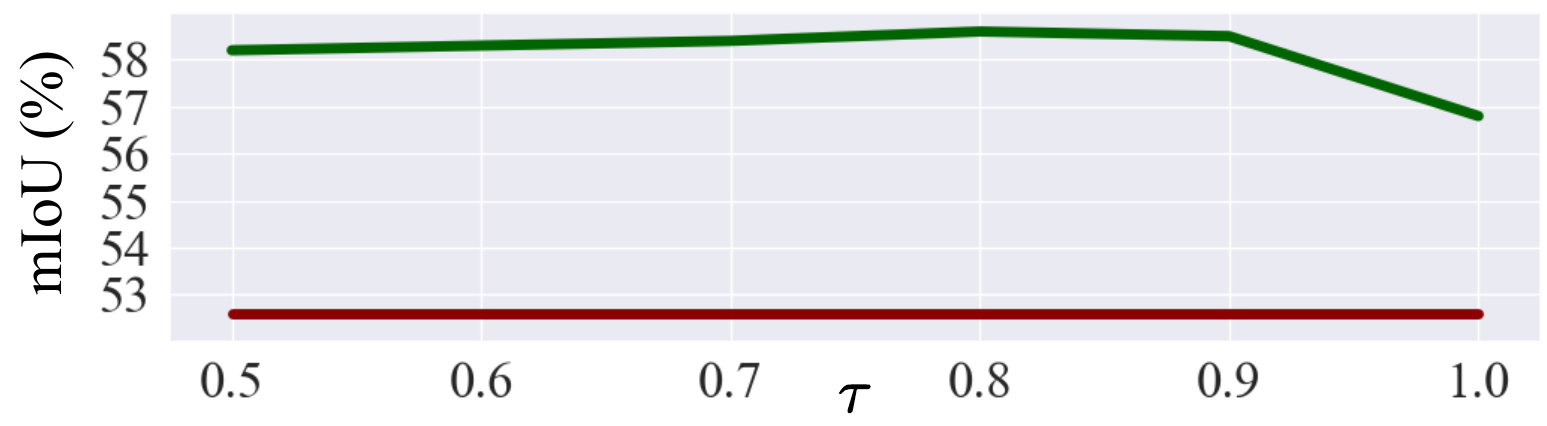}
\vspace{-0.5cm}
\caption{\textbf{Sensitivity of a hyperparameter $\tau$}. All mIoU values are calculated using the COCO training dataset. The red line is our baseline \cite{cha2023learning, ren2024grounded, ru2023token, Jo_2023_ICCV}.}
\label{fig:hyper}
\end{figure}

\vspace{-0.3cm}

\subsubsection{Hyperparameter of DHR} 
In our study, we evaluate the impact of the hyperparameter $\tau$ in Eq. \eqref{eq:wss} for WSS balancing, specifically after segmenting inter-class regions using USS features, as detailed in Eq. \eqref{eq:uss}. As illustrated in Figure \ref{fig:hyper}, setting $\tau$ to 1.0, which bypasses WSS balancing, results in only a slight improvement. This result is attributed to increased in incorrect predictions within inter-class areas when WSS features are omitted. Adjusting $\tau$ between 0.5 and 0.9 demonstrates minimal impact on segmentation accuracy, highlighting our DHR resilience to variations in $\tau$.

\vspace{-0.5cm}

\subsection{Qualitative Results} 
\begin{figure}[!t]
\centering
\includegraphics[width=0.9\linewidth]{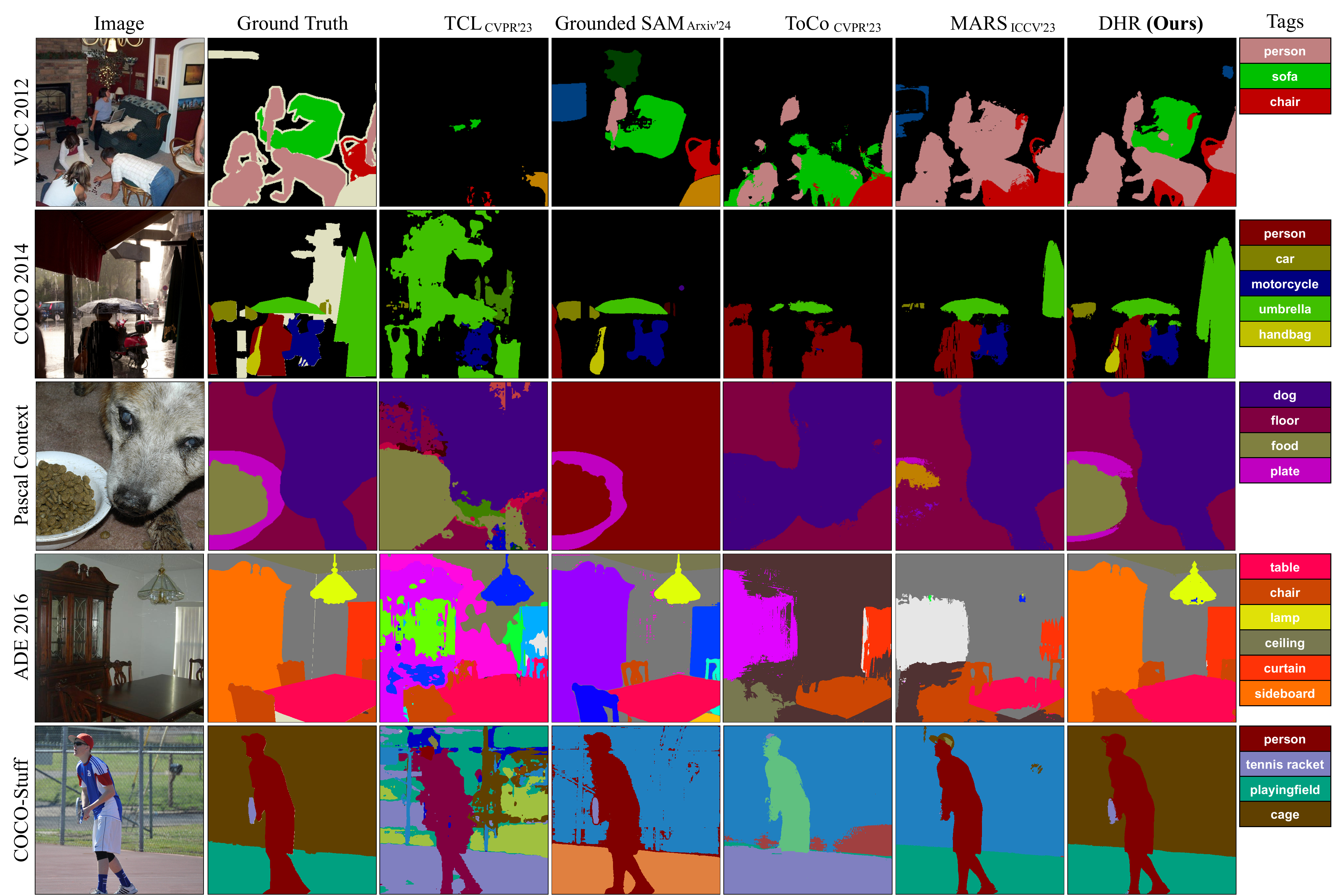}
\vspace{-0.35cm}
\caption{Qualitative comparison with ours and recent models \cite{cha2023learning, ren2024grounded, ru2023token, Jo_2023_ICCV}.}
\label{fig:qualitative}
\end{figure}

\vspace{-0.1cm}
Figure \ref{fig:qualitative} compares our DHR's segmentation performance against leading WSS methodologies across five benchmark datasets. The results underscore DHR's enhanced capability in delineating semantic boundaries between adjacent classes, outperforming existing state-of-the-art approaches \cite{cha2023learning, ren2024grounded, ru2023token, Jo_2023_ICCV} using tag inputs. For further examples, refer to Appendix \ref{sec:qualitative_results}.

\vspace{-0.4cm}

\section{Conclusion}
\label{sec:conclusion}
\vspace{-0.3cm}
In this study, we present DHR, a novel propagation strategy employing a hierarchical integration of unsupervised and weakly-supervised features to address the issue of adjacent minor classes disappearing. This method has significantly enhanced the performance of leading WSS models, narrowing the gap with FSS by over 50\% across five segmentation benchmarks, promising significant impacts on WSS research and extending its use across various fields (\emph{e.g.}, robotics, scene understanding, and autonomous driving) where propagation strategies are essential, showcasing our method's broad utility. DHR distinguishes itself from major vision models like CLIP \cite{li2021contrastive} and Grounded SAM \cite{ren2024grounded} by producing accurate segmentation masks with fewer samples for specific classes. It has become a valuable resource in the industrial and medical fields, where labeling is costly. When used as a starting point for the latest interactive segmentation tools, such as SAM, DHR outperforms other technologies for providing initial seeds, underscoring its potential in WSS. Its efficiency with minimal annotations and limited datasets showcases its considerable promise in segmentation tasks, underlining the importance of WSS in scenarios with significant annotation obstacles.

\clearpage 

\appendix

\section{Method Details}
\label{sec:method_details}

\subsection{Class-level Average Pooling}
\label{sec:cap}
\begin{figure}[h] 
\centering
\includegraphics[width=0.90\linewidth]{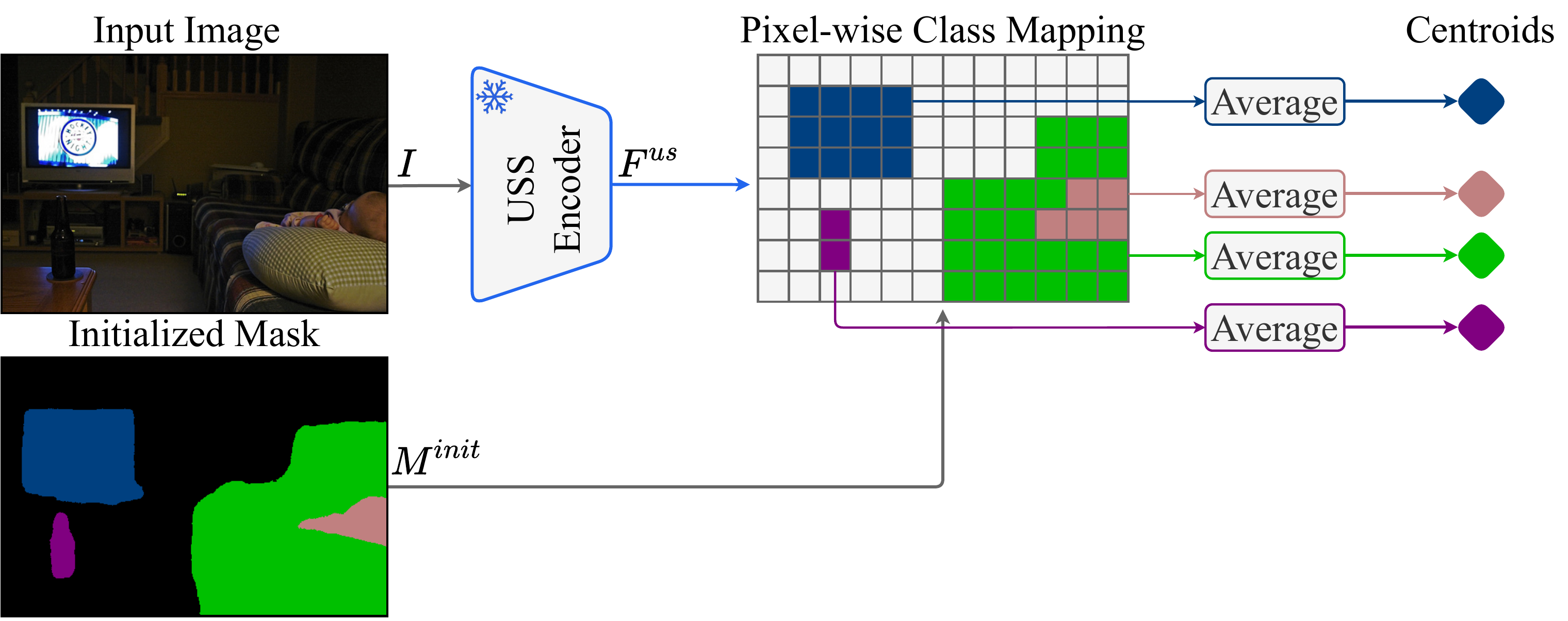}
\vspace{-0.3cm}
\caption{\textbf{Diagram of class-level average pooling.} Embedding vectors for each class are grouped based on the mask. Subsequently, class-specific centroids are computed as the average of those grouped vectors.}
\label{fig:cap}
\end{figure}

To derive class-specific centroids \(V^{us}\) from the WSS mask \(M^{init}\) as outlined in Eq. \eqref{eq:ot}, we modify a standard pooling technique, such as global average pooling, as demonstrated in Figure \ref{fig:cap}.

\vspace{-0.5cm}

\vspace{-0.5cm}

\begin{figure}[!b]
\centering
\includegraphics[width=0.90\linewidth]{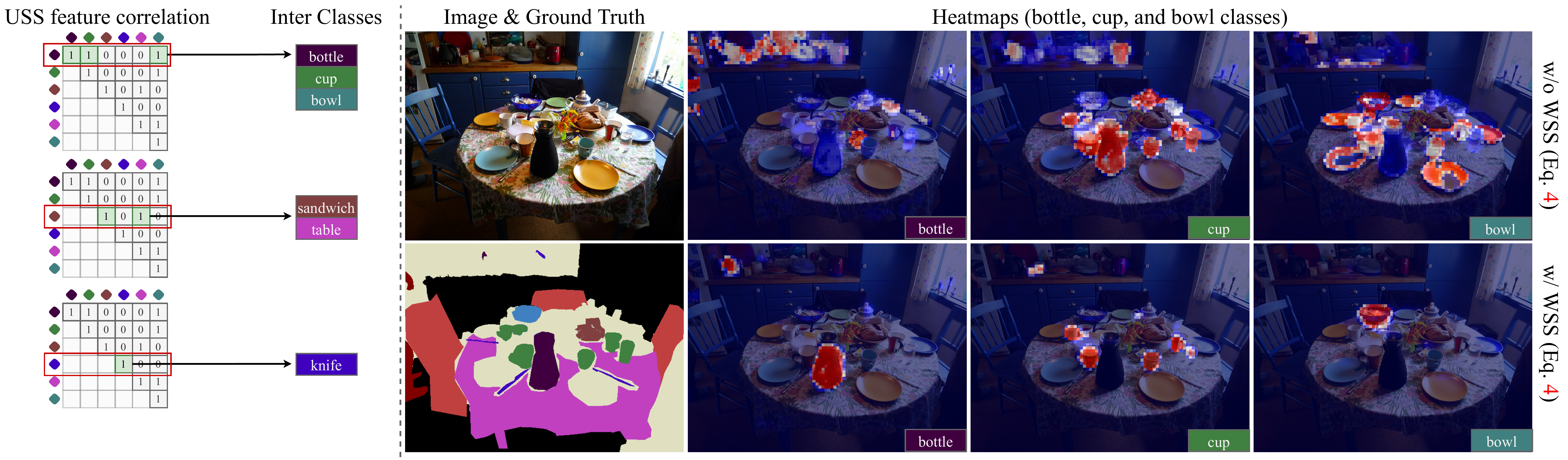}
\vspace{-0.3cm}
\caption{\textbf{Visualization of heatmaps with/without WSS balancing in Eq. \eqref{eq:wss}.}}
\label{fig:detailed_wss}
\end{figure}

\subsection{Details of USS Feature Correlation Matrix}
\label{sec:uss_feature_correlation}
Figure \ref{fig:detailed_wss} serves as an extensive elaboration of Figure \ref{fig:rebalancing}, further illustrating the capabilities of our Dual Features-Driven Hierarchical Rebalancing (DHR) method. The left side shows how DHR clusters all inter-class regions by leveraging a binarized USS feature correlation matrix \(\mathbbm{1}_{[{sim}(V^{us}, V^{us}) > \tau]}\). Following this automatic grouping, DHR employs WSS feature maps, as specified in Eq. \eqref{eq:wss}, to delineate these inter-class boundaries effectively. On the right, Figure \ref{fig:detailed_wss} contrasts heatmaps generated with and without applying WSS balancing. Applying DHR with WSS balancing yields precise heatmaps that distinguish between closely related inter-class regions, such as bottle, cup, and bowl, demonstrating our DHR's efficacy in enhancing segmentation accuracy in adjacent inter- and intra-class regions.



\subsection{Computational Complexity}
\label{sec:additional_discussion}

To quantify the computational overhead of applying our DHR, we investigate the training and testing times across five datasets, as detailed in Table \ref{tab:rebuttal_complexity}. The total training time with DHR increases by a factor of 1.8 (from 10 hours to 18 hours on VOC 2012 \cite{everingham2010pascal}). To mitigate the computational overhead introduced by techniques such as OT \cite{rachev1985monge}, we utilize 64 CPU cores in parallel, effectively reducing the impact on the overall training time. Importantly, despite the increased complexity during training, the testing time remains consistent with all baselines \cite{ahn2019weakly, wang2020self, lee2021anti, jo2022recurseed}. This consistency ensures that our refining steps are applied exclusively during the training phase, maintaining model deployment efficiency. 

\vspace{-0.5cm}

\begin{table}
\centering
  \caption{ 
  {Complexity comparison of our baseline (\emph{i.e.}, RSEPM \cite{jo2022recurseed}) with and without DHR. As all DHR steps in Sec. \ref{sec:dhr} are used only during training, the inference time is the same as all baselines \cite{ahn2019weakly, wang2020self, lee2021anti, jo2022recurseed}.}
  }
  \vspace{-0.3cm}
  \begin{scriptsize}
  \begin{tabular}{p{0.420\textwidth} | c | c}
    \toprule
    \textbf{\textcolor{blue}{(Dataset)}} Phase      & RSEPM \cite{jo2022recurseed} without DHR & RSEPM \cite{jo2022recurseed} with DHR \\
    \hline \hline
    \textbf{\textcolor{blue}{(VOC 2012)}} Total Training Time & 10 hours & 18 hours (\textbf{\textcolor{RED}{+8 hours}}) \\
    \textbf{\textcolor{blue}{(VOC 2012)}} Total Testing Time & 5 minutes & 5 minutes (\textbf{\textcolor{GREEN}{+0 minutes}}) \\
    \hline
    \textbf{\textcolor{blue}{(COCO 2014)}} Total Training Time & 19 hours & 36 hours (\textbf{\textcolor{RED}{+17 hours}}) \\
    \textbf{\textcolor{blue}{(COCO 2014)}} Total Testing Time & 30 minutes & 30 minutes (\textbf{\textcolor{GREEN}{+0 minutes}}) \\
    \hline
    \textbf{\textcolor{blue}{(Context)}} Total Training Time & 5 hours & 9 hours (\textbf{\textcolor{RED}{+4 hours}}) \\
    \textbf{\textcolor{blue}{(Context)}} Total Testing Time & 10 minutes & 10 minutes (\textbf{\textcolor{GREEN}{+0 minutes}}) \\
    \hline
    \textbf{\textcolor{blue}{(ADE 2016)}} Total Training Time & 11 hours & 20 hours (\textbf{\textcolor{RED}{+9 hours}}) \\
    \textbf{\textcolor{blue}{(ADE 2016)}} Total Testing Time & 15 minutes & 15 minutes (\textbf{\textcolor{GREEN}{+0 minutes}}) \\
    \hline
    \textbf{\textcolor{blue}{(Stuff)}} Total Training Time & 40 hours & 73 hours (\textbf{\textcolor{RED}{+33 hours}}) \\
    \textbf{\textcolor{blue}{(Stuff)}} Total Testing Time & 20 minutes & 20 minutes (\textbf{\textcolor{GREEN}{+0 minutes}}) \\
    \hline
    \bottomrule
  \end{tabular}
  \label{tab:rebuttal_complexity}
  \end{scriptsize}
  \vspace{-0.6cm}
\end{table}

\vspace{-0.5cm}


\section{Additional Quantitative Results}
\label{sec:additional_quantitative_results}

\vspace{-0.5cm}

\subsection{State-of-the-art Results with Other Architectures}
\label{sec:sota_architecture}

\vspace{-0.25cm}

In our commitment to a balanced evaluation, we compare our method against other WSS studies, specifically those built on ResNet architectures \cite{he2016deep}. Notably, recent WSS methods incorporate advanced supervision (\emph{e.g.}, CLIP \cite{li2021contrastive}) beyond image-level supervision and integrate cutting-edge decoders like Mask2Former \cite{cheng2022masked}. In response to this evolving landscape, we also adapt our DHR approach to compatibility with various backbone architectures \cite{liu2021swin} and decoders \cite{cheng2022masked,yang2024weakly} in Table \ref{tab:appendix_performance}, mirroring these contemporary configurations \cite{yang2024foundation}. Remarkably, using ResNet-101 and Swin-L backbones, our results are 79.8\% and 82.1\% on the VOC test set, respectively. These figures represent 98.5\% and 95.3\% of the fully-supervised upper bound performance at 86.1\%, significantly closing the gap between WSS and FSS. 

\vspace{-0.25cm}

\subsection{Per-class Performance Analysis}
\label{sec:per-class_analysis}

\vspace{-0.25cm}

Tables \ref{tab:voc_val_detail} and \ref{tab:voc_test_detail} detail per-class segmentation outcomes for the PASCAL VOC dataset. Addressing the issue of minor-class disappearance in adjacent pixels leads to enhancements across all classes rather than improvements confined to specific categories.

\subsection{Consistent Improvements on Other Datasets}
\label{sec:consistent_improvements}
In addition to the key component experiments conducted on the COCO dataset \cite{lin2014microsoft}, as shown in Table \ref{tab:ablation}, we extend our analysis to include detailed experimental results for the Context and ADE datasets \cite{mottaghi2014role, zhou2019semantic} (see Tables \ref{tab:ablation_context} and \ref{tab:ablation_ade}). These further analyses show that our DHR method enhances performance significantly, achieving up to a 17.9\% improvement on two datasets, which feature adjacent class scenarios.

\vspace{-0.5cm}

\begin{table}
  \centering
  \caption{Detailed analysis of each component on the Pascal Context training dataset.}
  \vspace{-0.3cm}
  \begin{scriptsize}

\begin{tabular}{>{\centering}p{0.20\textwidth} >{\centering}p{0.20\textwidth} | >{\centering}p{0.20\textwidth} >{\centering}p{0.20\textwidth} | p{0.15\textwidth}}
    \toprule
    \multicolumn{2}{c|}{\textbf{OT-based Seed Initialization}} & \multicolumn{2}{c|}{\textbf{DHR}} & \multirow{2}{*}{mIoU} \\
    Optimal Transport & Refinement & USS Balancing & WSS Balancing &  \\
    \midrule
    \xmark & \xmark & \xmark & \xmark & 51.3 \\
    \cmark & \xmark & \xmark & \xmark & 52.7 (+1.4) \\
    \cmark & \cmark & \xmark & \xmark & 53.6 (+2.3) \\
    \midrule
    \cmark & \cmark & \cmark & \xmark & 61.1 (+9.8) \\
    \cmark & \cmark & \xmark & \cmark & 54.6 (+3.3) \\
    \rowcolor[HTML]{F4CCCC} 
    \cmark & \cmark & \cmark & \cmark & \textbf{64.9 (+13.6)} \\
    \bottomrule
\end{tabular}

  \label{tab:ablation_context}
  \end{scriptsize}
\end{table}

\vspace{-1.25cm}

\begin{table}
  \centering
  \caption{Detailed analysis of each component on the ADE training dataset.}
  \vspace{-0.3cm}
  \begin{scriptsize}

\begin{tabular}{>{\centering}p{0.20\textwidth} >{\centering}p{0.20\textwidth} | >{\centering}p{0.20\textwidth} >{\centering}p{0.20\textwidth} | p{0.15\textwidth}}
    \toprule
    \multicolumn{2}{c|}{\textbf{OT-based Seed Initialization}} & \multicolumn{2}{c|}{\textbf{DHR}} & \multirow{2}{*}{mIoU} \\
    Optimal Transport & Refinement & USS Balancing & WSS Balancing &  \\
    \midrule
    \xmark & \xmark & \xmark & \xmark & 30.2 \\
    \cmark & \xmark & \xmark & \xmark & 32.0 (+1.8) \\
    \cmark & \cmark & \xmark & \xmark & 32.4 (+2.2) \\
    \midrule
    \cmark & \cmark & \cmark & \xmark & 44.7 (+14.5) \\
    \cmark & \cmark & \xmark & \cmark & 38.1 (+7.9) \\
    \rowcolor[HTML]{F4CCCC} 
    \cmark & \cmark & \cmark & \cmark & \textbf{48.1 (+17.9)} \\
    \bottomrule
\end{tabular}

  \label{tab:ablation_ade}
  \end{scriptsize}
\end{table}

\vspace{-1.0cm}

\section{Additional Qualitative Results}
\label{sec:additional_qualitative_results}

\subsection{Model-agnostic Improvements}
\label{sec:qualitative_improvements}

\begin{figure}[!t]
\centering
\includegraphics[width=\linewidth]{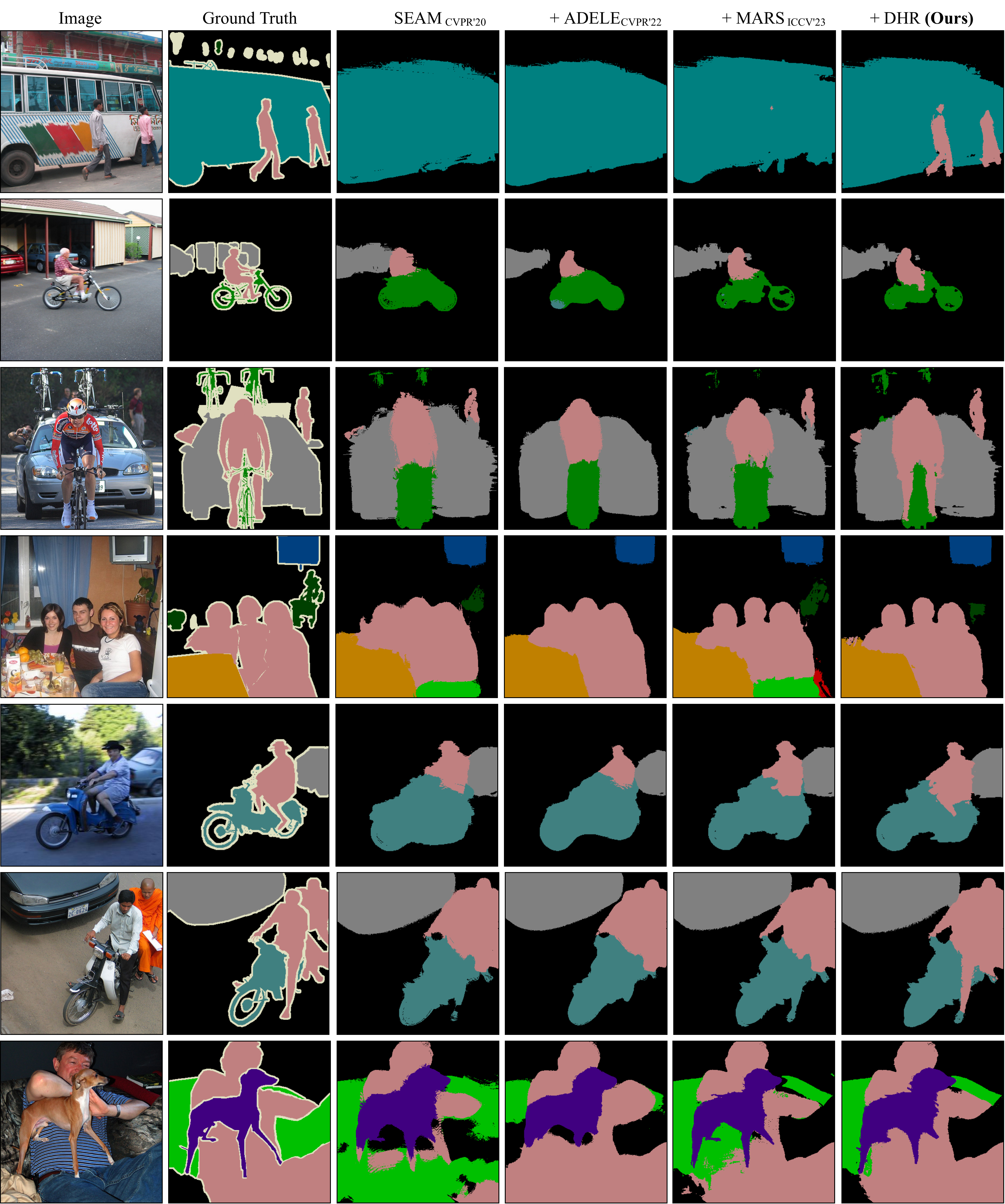}
\caption{Qualitative results with ours and other model-agnostic models \cite{liu2022adaptive, Jo_2023_ICCV}.}
\label{fig:qualitative-seam}
\end{figure}

\begin{figure}[!t]
\centering
\includegraphics[width=\linewidth]{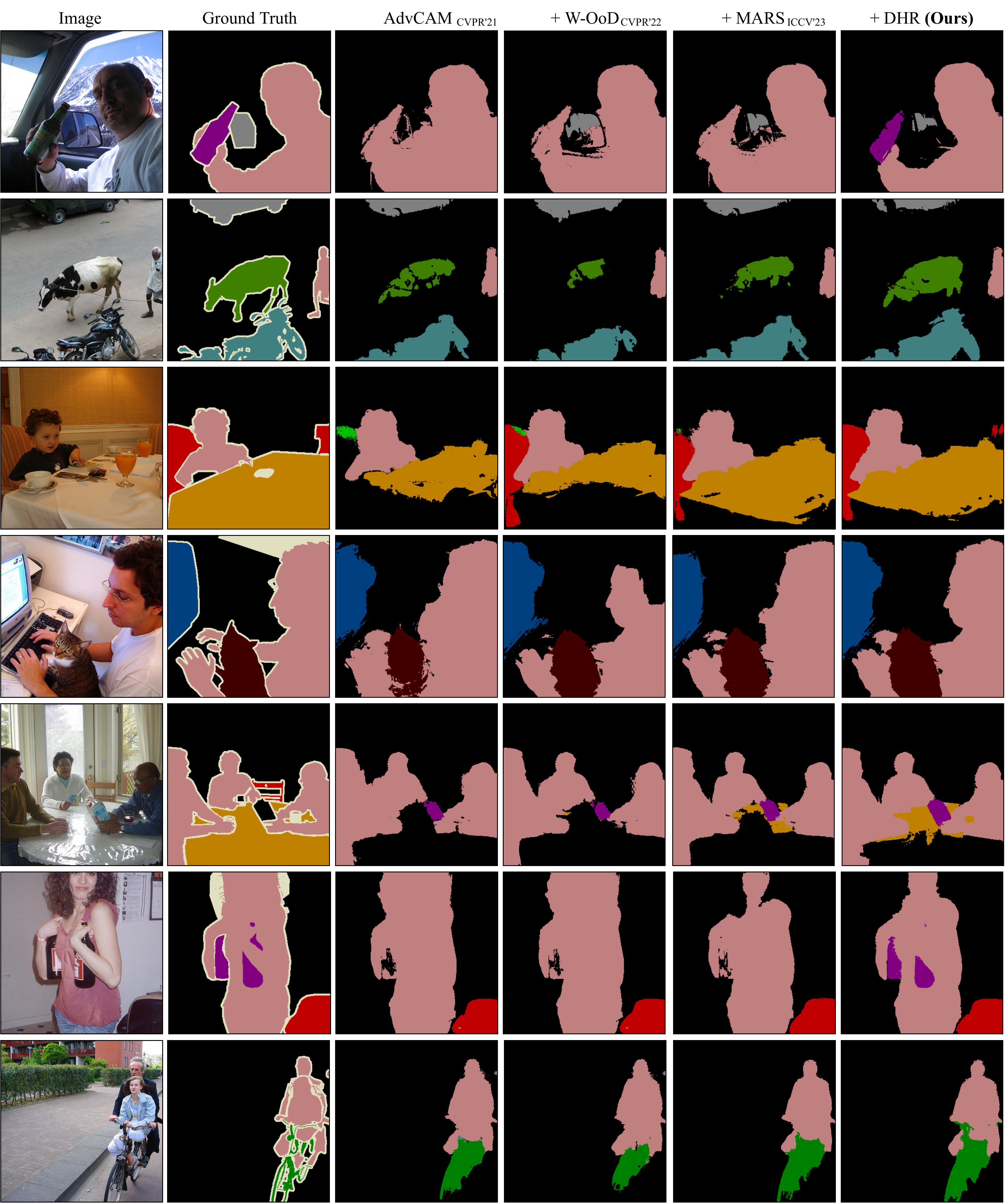}
\caption{Qualitative results with ours and other model-agnostic models \cite{lee2021anti, Jo_2023_ICCV}.}
\label{fig:qualitative-advcam}
\end{figure}

Figures \ref{fig:qualitative-seam} and \ref{fig:qualitative-advcam} provide a qualitative comparison between our DHR method, two baseline models \cite{wang2020self,lee2021anti}, and various model-agnostic approaches \cite{liu2022adaptive, lee2022weakly, Jo_2023_ICCV}. Demonstrating its robustness, DHR excels in segmenting diverse objects and managing scenarios with multiple classes. It particularly shines in accurately segmenting adjacent minor-class regions (\emph{e.g.}, person and bicycle), where it significantly outperforms other WSS methods \cite{liu2022adaptive, lee2022weakly, Jo_2023_ICCV} by recovering classes that are often missed or overlooked, thereby ensuring comprehensive and satisfactory segmentation outcomes on five benchmarks.

\subsection{Qualitative Segmentation Examples}
\label{sec:qualitative_results}
\begin{figure}[!t]
\centering
\includegraphics[width=\linewidth]{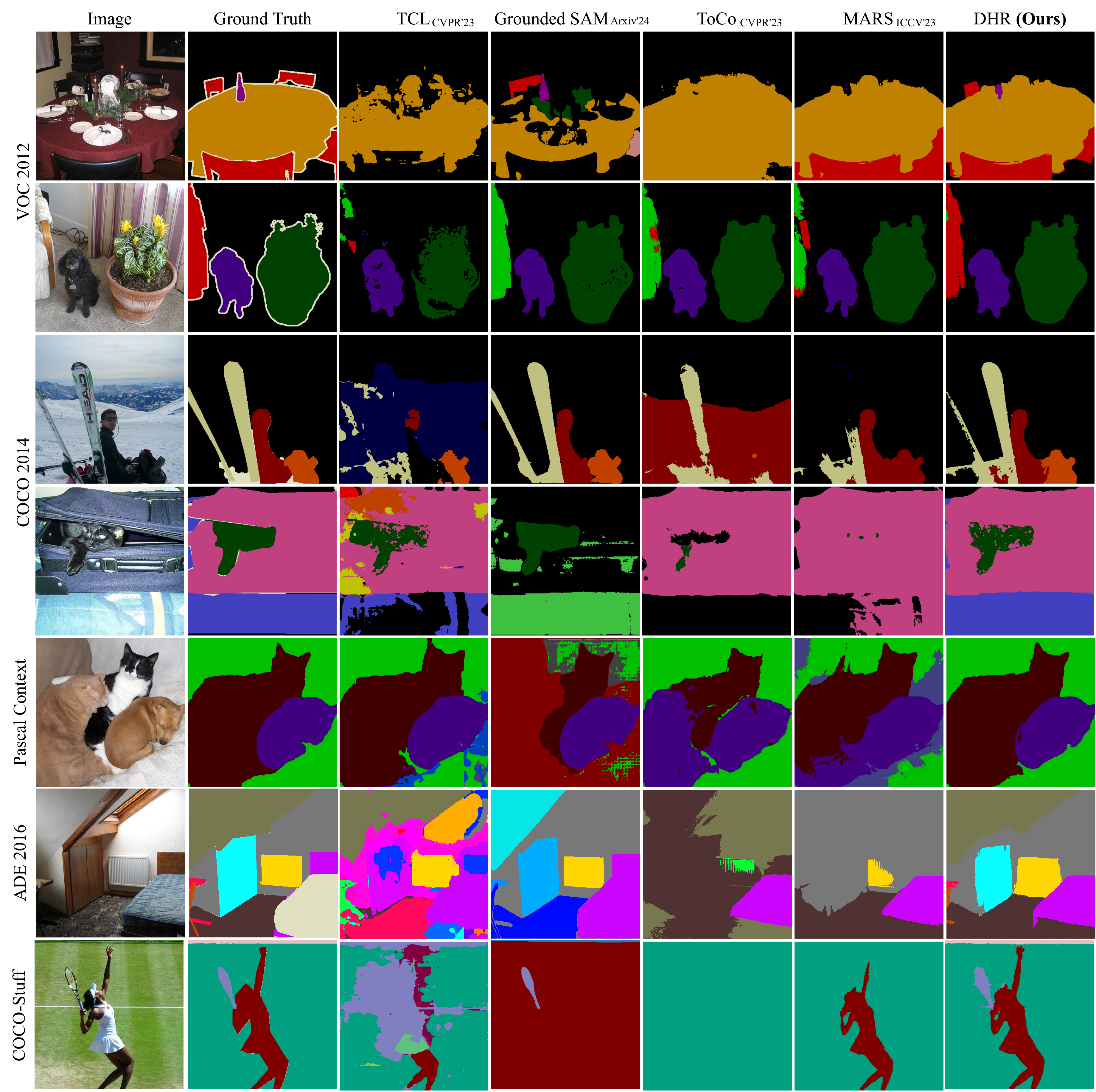}
\caption{Qualitative results with ours and recent state-of-the-art methods \cite{cha2023learning, ren2024grounded, ru2023token, Jo_2023_ICCV}.}
\label{fig:qualitative-addition}
\end{figure}

When compared against recent open-vocabulary and WSS models \cite{cha2023learning, ren2024grounded, ru2023token, Jo_2023_ICCV} across five benchmark datasets, our DHR exhibits outstanding performance both qualitatively and quantitatively, surpassing previous state-of-the-art methods (see Figure \ref{fig:qualitative-addition}). This comparison underscores the efficacy of DHR in handling real-world datasets characterized by multiple labels and intricate inter-/intra-class relationships, highlighting its potential to advance the field of semantic segmentation significantly.

\clearpage

\begin{table}[t]
    \centering
    \caption{State-of-the-art results compared to WSS methods with other backbones.}
  \vspace{-0.3cm}
\resizebox{\textwidth}{!}{
    
    \begin{tabular}{p{0.60\textwidth}cccccccc}
    \toprule
    \multicolumn{1}{c}{} &  &  & \multicolumn{2}{c}{VOC} & COCO & Context & ADE & Stuff \\  
    \multicolumn{1}{l}{\multirow{-2}{*}{Method}} & \multirow{-2}{*}{Backbone} & \multirow{-2}{*}{Supervision} & val & test & val & val & val & val \\
    \midrule
    \multicolumn{9}{l}{\textbf{WSS based on CNN architectures:}} \\
    DSRG {\tiny CVPR'18} \cite{huang2018weakly} & ResNet-101 & $\mathcal{I}$+$\mathcal{A}$ & 61.4 & 63.2 & 26.0 & - & - & - \\
    PSA {\tiny CVPR'18} \cite{ahn2018learning} & Wide-ResNet-38 & $\mathcal{I}$ & 61.7 & 63.7 & - & - & - & - \\
    SSSS {\tiny CVPR'20} \cite{araslanov2020single} & Wide-ResNet-38 & $\mathcal{I}$ & 62.7 & 64.3 & - & - & - & - \\
    IRNet {\tiny CVPR'19} \cite{ahn2019weakly} & ResNet-50 & $\mathcal{I}$ & 63.5 & 64.8 & - & - & - & - \\
    ICD {\tiny CVPR'20} \cite{fan2020learning} & ResNet-101 & $\mathcal{I}$ & 64.1 & 64.3 & - & - & - & - \\
    SEAM {\tiny CVPR'20} \cite{wang2020self} & Wide-ResNet-38 & $\mathcal{I}$ & 64.5 & 65.7 & 31.9 & - & - & - \\
    FickleNet {\tiny CVPR'19} \cite{lee2019ficklenet} & ResNet-101 & $\mathcal{I}$+$\mathcal{A}$ & 64.9 & 65.3 & -  & - & - & - \\
    RRM {\tiny AAAI'20} \cite{zhang2020reliability} & ResNet-101 & $\mathcal{I}$ & 66.3 & 65.5 & - & - & - & - \\
    RIB {\tiny NeurIPS'21} \cite{lee2021reducing} & ResNet-101 & $\mathcal{I}$ & 68.3 & 68.6 & 43.8 & - & - & - \\
    ReCAM {\tiny CVPR'22} \cite{chen2022class} & ResNet-101 & $\mathcal{I}$ & 68.5 & 68.4 & - & - & - & - \\
    AMR {\tiny AAAI'22} \cite{qin2022activation} & ResNet-101 & $\mathcal{I}$ & 68.8 & 69.1 & - & - & - & - \\
    URN {\tiny AAAI'22} \cite{li2022uncertainty} & ResNet-101 & $\mathcal{I}$ & 69.5 & 69.7 & 40.7 & - & - & - \\
    W-OoD {\tiny CVPR'22} \cite{lee2022weakly} & ResNet-101 & $\mathcal{I}$+$\mathcal{D}$ & 69.8 & 69.9 & -  & - & - & - \\
    EDAM {\tiny CVPR'21} \cite{wu2021embedded} & ResNet-101 & $\mathcal{I}$+$\mathcal{A}$ & 70.9 & 70.6 
    & -  & - & - & - \\
    EPS {\tiny CVPR'21} \cite{lee2021railroad} & ResNet-101 & $\mathcal{I}$+$\mathcal{A}$ & 70.9 & 70.8 
    & 35.7 & - & - & - \\
    SANCE {\tiny CVPR'22} \cite{li2022towards} & ResNet-101 & $\mathcal{I}$ & 70.9 & 72.2 & 44.7 & - & - & - \\
    DRS {\tiny AAAI'21} \cite{kim2021discriminative} & ResNet-101 & $\mathcal{I}$+$\mathcal{A}$ & 71.2 & 71.4 & - & - & - & - \\
    MCTformer {\tiny CVPR'22} \cite{xu2022multi} & Wide-ResNet-38 & $\mathcal{I}$ & 71.9 & 71.6 & 42.0 & - & - & - \\
    L2G {\tiny CVPR'22} \cite{jiang2022l2g} & ResNet-101 & $\mathcal{I}$+$\mathcal{A}$ & 72.1 & 71.7 & 44.2  & - & - & - \\
    RCA {\tiny CVPR'22} \cite{zhou2022regional} & ResNet-101 & $\mathcal{I}$+$\mathcal{A}$ & 72.2 & 72.8 & 36.8*  & - & - & - \\
    PPC {\tiny CVPR'22} \cite{du2022weakly} & ResNet-101 & $\mathcal{I}$+$\mathcal{A}$ & 72.6 & 73.6 & - & - & - & - \\
    SAS {\tiny AAAI'23} \cite{kim2023semantic} & ResNet-101 & $\mathcal{I}$ & 69.5 & 70.1 & 44.8 & - & - & - \\
    Jiang et al {\tiny arXiv'23} \cite{jiang2023segment} & ResNet-101 & $\mathcal{I}$+$\mathcal{S}$ & 71.1 & 72.2 & - & - & - & - \\
    ACR {\tiny CVPR'23} \cite{kweon2023weakly} & Wide-ResNet-38 &  $\mathcal{I}$ & 71.9 & 71.9 & 45.3 & - & - & - \\
    BECO {\tiny CVPR'23} \cite{Rong_2023_CVPR} & ResNet-101 &  $\mathcal{I}$ & 72.1 & 71.8 & - & - & - & - \\
    MMSCT {\tiny CVPR'23} \cite{xu2023learning} & Wide-ResNet-38 & $\mathcal{I}$+$\mathcal{C}$ & 72.2 & 72.2 & 45.9 & - & - & - \\
    QA-CLIMS {\tiny MM'23} \cite{deng2023qa-clims} & ResNet-101 & $\mathcal{I}$+$\mathcal{L}$ & 72.4 & 72.3 & 43.2 & - & - & - \\
    OCR {\tiny CVPR'23} \cite{cheng2023out} & Wide-ResNet-38 & $\mathcal{I}$ & 72.7 & 72.0 & 42.5 & - & - & - \\
    CLIP-ES {\tiny CVPR'23} \cite{Lin_2023_CVPR} & ResNet-101 & $\mathcal{I}$+$\mathcal{C}$ & 73.8 & 73.9 & 45.4 & - & - & - \\
    RSEPM {\tiny arXiv'22} \cite{jo2022recurseed} & ResNet-101 & $\mathcal{I}$ & 74.4 & 73.6 & 46.4 & - & - & - \\
    CoSA {\tiny arXiv'24} \cite{zhu2023weaktr} & ResNet-101 &  $\mathcal{I}$ & 76.5 & 75.3 & 50.9 & - & - & - \\
    FMA-WSSS {\tiny WACV'24} \cite{yang2024foundation} & ResNet-101 & $\mathcal{I}$+$\mathcal{C}$+$\mathcal{S}$ & 77.3 & 76.7 & 48.6 & - & - & - \\
    MARS {\tiny ICCV'23} \cite{Jo_2023_ICCV} & ResNet-101 &  $\mathcal{I}$ & 77.7 & 77.2 & 49.4 & 39.8* & 22.0* & 35.7* \\
    \rowcolor[HTML]{F4CCCC} 
    \multicolumn{1}{l}{\cellcolor[HTML]{F4CCCC}\textbf{DHR (Ours)}} & ResNet-101 &  $\mathcal{I}$ & \textbf{79.6}  & \textbf{79.8} & \textbf{53.9} & \textbf{49.0} & \textbf{32.9} & \textbf{37.4} \\
   \textcolor{gray}{ Upper Bound (DeepLabv3+ {\tiny CVPR'18} \cite{chen2018encoder})} &\textcolor{gray}{ ResNet-101} &\textcolor{gray}{  $\mathcal{M}$} &\textcolor{gray}{ 80.6*} &\textcolor{gray}{ 81.0*} &\textcolor{gray}{ 61.8*} &\textcolor{gray}{ 54.6*} &\textcolor{gray}{ 45.3*} &\textcolor{gray}{ 44.2*} \\
    \midrule
    \multicolumn{9}{l}{\textbf{WSS based on Transformer architectures:}} \\
    BECO {\tiny CVPR'23} \cite{Rong_2023_CVPR} & MiT-B2 &  $\mathcal{I}$ & 73.7 & 73.5 & 45.1 & - & - & - \\
    ToCo {\tiny CVPR'23} \cite{ru2023token} & ViT-B &  $\mathcal{I}$ & 71.1 & 72.2 & 42.3 & 25.0* & 10.5* & 14.2* \\
    WeakTr {\tiny arXiv'23} \cite{zhu2023weaktr} & DeiT-S &  $\mathcal{I}$ & 74.0 & 74.1 & 46.9 & - & - & - \\
    CoSA {\tiny arXiv'24} \cite{zhu2023weaktr} & Swin-B &  $\mathcal{I}$ & 81.4 & 78.4 & 53.7 & - & - & - \\
    FMA-WSSS {\tiny WACV'24} \cite{yang2024foundation} & Swin-L & $\mathcal{I}$+$\mathcal{C}$+$\mathcal{S}$ & 82.6 & 81.6 & 55.4 & - & - & - \\
    \rowcolor[HTML]{F4CCCC} 
    \cellcolor[HTML]{F4CCCC}\textbf{DHR (Ours)} & Swin-L &  $\mathcal{I}$ & \textbf{82.3} & \textbf{82.3} & \textbf{56.8} & \textbf{53.6} & \textbf{36.9} & \textbf{41.1} \\
   \textcolor{gray}{Upper Bound (Mask2Former {\tiny CVPR'22} \cite{cheng2022masked})} &\textcolor{gray}{ Swin-L} & \textcolor{gray}{ $\mathcal{M}$} & \textcolor{gray}{86.0} & \textcolor{gray}{86.1} & \textcolor{gray}{66.7} & \textcolor{gray}{64.3*} & \textcolor{gray}{55.5*} & \textcolor{gray}{50.6*} \\
   \midrule
    \multicolumn{9}{l}{\textbf{Open-vocabulary Segmentation Models:}} \\
    MaskCLIP {\tiny ECCV'22} \cite{zhou2022extract} & ViT-B & $\mathcal{C}$+$\mathcal{T}$  & 29.3 & - & 15.5 & 21.1 & 10.8 & 14.7 \\
    TCL {\tiny CVPR'23} \cite{cha2023learning} & ViT-B & $\mathcal{C}$+$\mathcal{T}$  & 55.0 & - & 33.2 & 33.8 & 15.6 & 22.4 \\
    Ferret {\tiny arXiv'23} \cite{you2023ferret} w/ SAM \cite{ke2024segment} & ViT-H & $\mathcal{B}$+$\mathcal{T}$+$\mathcal{S}$+$\mathcal{L}$ & 54.7 & - & 27.7 & 22.4 & 7.6 & 12.6 \\
    Grounded SAM {\tiny arXiv'24} \cite{ren2024grounded} & Swin-B &$\mathcal{B}$+$\mathcal{T}$+$\mathcal{S}$ & 46.3 & - & 35.7 & 28.1 & 4.8 & 18.8 \\
    \bottomrule
*: we reproduce all results for a fair comparison & & & & & & & & \\ 
$\mathcal{I}$: image-level supervision & \multicolumn{3}{l}{$\mathcal{T}$: text supervision (image-text pairs)} & & & \multicolumn{2}{l}{$\mathcal{A}$: saliency \cite{hou2017deeply}} & \\
$\mathcal{L}$: language supervision (\emph{e.g.}, LLM \cite{you2023ferret}) & \multicolumn{2}{l}{$\mathcal{M}$: mask supervision}& \multicolumn{2}{l}{$\mathcal{S}$: SAM \cite{ke2024segment}} & & \multicolumn{2}{l}{$\mathcal{C}$: CLIP \cite{radford2021learning}} \\ $\mathcal{B}$: box supervision & \multicolumn{2}{l}{$\mathcal{D}$: using the external dataset \cite{lee2022weakly}} \\
    
\end{tabular}
}
  \label{tab:appendix_performance}
\end{table}


\clearpage

\begin{table}[!t]
  \centering
  \caption{
    Per-class performance comparison with WSS methods in terms of IoUs (\%) on the PASCAL VOC 2012 validation set.
  }
  \begin{scriptsize}
  \resizebox{\textwidth}{!}{
  \begin{tabular}{
    p{0.25\textwidth} 
    p{0.043\textwidth} p{0.043\textwidth} p{0.043\textwidth} p{0.043\textwidth} p{0.043\textwidth} 
    p{0.043\textwidth} p{0.043\textwidth} p{0.043\textwidth} p{0.043\textwidth} p{0.043\textwidth} 
    p{0.043\textwidth} p{0.043\textwidth} p{0.043\textwidth} p{0.043\textwidth} p{0.043\textwidth} 
    p{0.043\textwidth} p{0.043\textwidth} p{0.043\textwidth} p{0.043\textwidth} p{0.043\textwidth} p{0.043\textwidth} 
    c }
    \toprule
    Method & \rotatebox[origin=l]{90}{bkg} & \rotatebox[origin=l]{90}{aero} & \rotatebox[origin=l]{90}{bike} & \rotatebox[origin=l]{90}{bird} & \rotatebox[origin=l]{90}{boat} & \rotatebox[origin=l]{90}{bottle} & \rotatebox[origin=l]{90}{bus} & \rotatebox[origin=l]{90}{car} & \rotatebox[origin=l]{90}{cat} & \rotatebox[origin=l]{90}{chair} & \rotatebox[origin=l]{90}{cow} & \rotatebox[origin=l]{90}{table} & \rotatebox[origin=l]{90}{dog} & \rotatebox[origin=l]{90}{horse} & \rotatebox[origin=l]{90}{mbk} & \rotatebox[origin=l]{90}{person} & \rotatebox[origin=l]{90}{plant} & \rotatebox[origin=l]{90}{sheep} & \rotatebox[origin=l]{90}{sofa} & \rotatebox[origin=l]{90}{train} & \rotatebox[origin=l]{90}{tv}  & mIoU  \\
    \hline \hline
    EM {\tiny ICCV'15} \cite{papandreou2015weakly} & 67.2 & 29.2 & 17.6 & 28.6 & 22.2 & 29.6 & 47.0 & 44.0 & 44.2 & 14.6 & 35.1 & 24.9 & 41.0 & 34.8 & 41.6 & 32.1 & 24.8 & 37.4 & 24.0 & 38.1 & 31.6 & 33.8 \\
    MIL-LSE {\tiny CVPR'15} \cite{pinheiro2015image} & 79.6 & 50.2 & 21.6 & 40.9 & 34.9 & 40.5 & 45.9 & 51.5 & 60.6 & 12.6 & 51.2 & 11.6 & 56.8 & 52.9 & 44.8 & 42.7 & 31.2 & 55.4 & 21.5 & 38.8 & 36.9 & 42.0 \\
    SEC {\tiny ECCV'16} \cite{kolesnikov2016seed} & 82.4 & 62.9 & 26.4 & 61.6 & 27.6 & 38.1 & 66.6 & 62.7 & 75.2 & 22.1 & 53.5 & 28.3 & 65.8 & 57.8 & 62.3 & 52.5 & 32.5 & 62.6 & 32.1 & 45.4 & 45.3 & 50.7 \\
    TransferNet {\tiny CVPR'16} \cite{hong2016learning} & 85.3 & 68.5 & 26.4 & 69.8 & 36.7 & 49.1 & 68.4 & 55.8 & 77.3 & 6.2 & 75.2 & 14.3 & 69.8 & 71.5 & 61.1 & 31.9 & 25.5 & 74.6 & 33.8 & 49.6 & 43.7 & 52.1 \\
    CRF-RNN {\tiny CVPR'17} \cite{roy2017combining} & 85.8 & 65.2 & 29.4 & 63.8 & 31.2 & 37.2 & 69.6 & 64.3 & 76.2 & 21.4 & 56.3 & 29.8 & 68.2 & 60.6 & 66.2 & 55.8 & 30.8 & 66.1 & 34.9 & 48.8 & 47.1 & 52.8 \\
    WebCrawl {\tiny CVPR'17} \cite{hong2017weakly} & 87.0 & 69.3 & 32.2 & 70.2 & 31.2 & 58.4 & 73.6 & 68.5 & 76.5 & 26.8 & 63.8 & 29.1 & 73.5 & 69.5 & 66.5 & 70.4 & 46.8 & 72.1 & 27.3 & 57.4 & 50.2 & 58.1 \\
    CIAN {\tiny AAAI'20} \cite{fan2020cian} & 88.2 & 79.5 & 32.6 & 75.7 & 56.8 & 72.1 & 85.3 & 72.9 & 81.7 & 27.6 & 73.3 & 39.8 & 76.4 & 77.0 & 74.9 & 66.8 & 46.6 & 81.0 & 29.1 & 60.4 & 53.3 & 64.3 \\
    SSDD {\tiny ICCV'19} \cite{shimoda2019self} & 89.0 & 62.5 & 28.9 & 83.7 & 52.9 & 59.5 & 77.6 & 73.7 & 87.0 & 34.0 & 83.7 & 47.6 & 84.1 & 77.0 & 73.9 & 69.6 & 29.8 & 84.0 & 43.2 & 68.0 & 53.4 & 64.9 \\
    PSA {\tiny CVPR'18} \cite{ahn2018learning} & 87.6 & 76.7 & 33.9 & 74.5 & 58.5 & 61.7 & 75.9 & 72.9 & 78.6 & 18.8 & 70.8 & 14.1 & 68.7 & 69.6 & 69.5 & 71.3 & 41.5 & 66.5 & 16.4 & 70.2 & 48.7 & 59.4 \\
    FickleNet {\tiny CVPR'19} \cite{lee2019ficklenet} & 89.5 & 76.6 & 32.6 & 74.6 & 51.5 & 71.1 & 83.4 & 74.4 & 83.6 & 24.1 & 73.4 & 47.4 & 78.2 & 74.0 & 68.8 & 73.2 & 47.8 & 79.9 & 37.0 & 57.3 & \textbf{64.6} & 64.9 \\
    RRM {\tiny AAAI'20} \cite{zhang2020reliability} & 87.9 & 75.9 & 31.7 & 78.3 & 54.6 & 62.2 & 80.5 & 73.7 & 71.2 & 30.5 & 67.4 & 40.9 & 71.8 & 66.2 & 70.3 & 72.6 & 49.0 & 70.7 & 38.4 & 62.7 & 58.4 & 62.6 \\
    SSSS {\tiny CVPR'20} \cite{araslanov2020single} & 88.7 & 70.4 & 35.1 & 75.7 & 51.9 & 65.8 & 71.9 & 64.2 & 81.1 & 30.8 & 73.3 & 28.1 & 81.6 & 69.1 & 62.6 & 74.8 & 48.6 & 71.0 & 40.1 & 68.5 & 64.3 & 62.7 \\
    SEAM {\tiny CVPR'20} \cite{wang2020self} & 88.8 & 68.5 & 33.3 & 85.7 & 40.4 & 67.3 & 78.9 & 76.3 & 81.9 & 29.1 & 75.5 & 48.1 & 79.9 & 73.8 & 71.4 & 75.2 & 48.9 & 79.8 & 40.9 & 58.2 & 53.0 & 64.5 \\
    AdvCAM {\tiny CVPR'21} \cite{lee2021anti} & 90.0 & 79.8 & 34.1 & 82.6 & 63.3 & 70.5 & 89.4 & 76.0 & 87.3 & 31.4 & 81.3 & 33.1 & 82.5 & 80.8 & 74.0 & 72.9 & 50.3 & 82.3 & 42.2 & 74.1 & 52.9 & 68.1 \\
    CPN {\tiny ICCV'21} \cite{zhang2021complementary} & 89.9 & 75.0 & 32.9 & 87.8 & 60.9 & 69.4 & 87.7 & 79.4 & 88.9 & 28.0 & 80.9 & 34.8 & 83.4 & 79.6 & 74.6 & 66.9 & 56.4 & 82.6 & 44.9 & 73.1 & 45.7 & 67.8 \\
    RIB {\tiny NeurIPS'21} \cite{lee2021reducing} & 90.3 & 76.2 & 33.7 & 82.5 & 64.9 & 73.1 & 88.4 & 78.6 & 88.7 & 32.3 & 80.1 & 37.5 & 83.6 & 79.7 & 75.8 & 71.8 & 47.5 & 84.3 & 44.6 & 65.9 & 54.9 & 68.3 \\
    AMN {\tiny CVPR'22} \cite{lee2022threshold} & 90.6 & 79.0 & 33.5 & 83.5 & 60.5 & 74.9 & 90.0 & 81.3 & 86.6 & 30.6 & 80.9 & 53.8 & 80.2 & 79.6 & 74.6 & 75.5 & 54.7 & 83.5 & 46.1 & 63.1 & 57.5 & 69.5 \\
    ADELE {\tiny CVPR'22} \cite{liu2022adaptive} & 91.1 & 77.6 & 33.0 & 88.9 & 67.1 & 71.7 & 88.8 & 82.5 & 89.0 & 26.6 & 83.8 & 44.6 & 84.4 & 77.8 & 74.8 & 78.5 & 43.8 & 84.8 & 44.6 & 56.1 & 65.3 & 69.3 \\
    W-OoD {\tiny CVPR'22} \cite{lee2022weakly} & 91.2 & 80.1 & 34.0 & 82.5 & 68.5 & 72.9 & 90.3 & 80.8 & 89.3 & 32.3 & 78.9 & 31.1 & 83.6 & 79.2 & 75.4 & 74.4 & 58.0 & 81.9 & 45.2 & 81.3 & 54.8 & 69.8 \\
    RCA {\tiny CVPR'22} \cite{zhou2022regional} & 91.8 & 88.4 & 39.1 & 85.1 & 69.0 & 75.7 & 86.6 & 82.3 & 89.1 & 28.1 & 81.9 & 37.9 & 85.9 & 79.4 & 82.1 & 78.6 & 47.7 & 84.4 & 34.9 & 75.4 & 58.6 & 70.6 \\
    SANCE {\tiny CVPR'22} \cite{li2022towards} & 91.4 & 78.4 & 33.0 & 87.6 & 61.9 & \textbf{79.6} & 90.6 & 82.0 & 92.4 & 33.3 & 76.9 & 59.7 & 86.4 & 78.0 & 76.9 & 77.7 & 61.1 & 79.4 & 47.5 & 62.1 & 53.3 & 70.9 \\
    MCTformer {\tiny CVPR'22} \cite{xu2022multi} & 91.9 & 78.3 & 39.5 & {89.9} & 55.9 & 76.7 & 81.8 & 79.0 & 90.7 & 32.6 & 87.1 & 57.2 & 87.0 & 84.6 & 77.4 & 79.2 & 55.1 & 89.2 & 47.2 & 70.4 & 58.8 & 71.9 \\
    RSEPM {\tiny arXiv'22} \cite{jo2022recurseed} & 92.2 & 88.4 & 35.4 & 87.9 & 63.8 & 79.5 & {93.0} & 84.5 & 92.7 & 39.0 & 90.5 & 54.5 & 90.6 & 87.5 & 83.0 & 84.0 & 61.1 & 85.6 & 52.1 & 56.2 & 60.2 & 74.4 \\
    MARS {\tiny ICCV'23} & {94.1} & {89.3} & \textbf{42.0} & 88.8 & {72.9} & 79.5 & 92.7 & \textbf{86.2} & {94.2} & {40.3} & {91.4} & 58.8 & 91.1 & 88.9 & 81.9 & 84.6 & 63.6 & 91.7 & 56.7 & 85.3 & 57.3 & 77.7 \\
    \rowcolor[HTML]{F4CCCC}
    \textbf{DHR (ResNet-101)} & \textbf{94.5} & \textbf{91.2} & 40.9 & \textbf{92.3} & \textbf{78.1} & 77.4 & \textbf{93.4} & \textbf{86.2} & \textbf{94.4} & \textbf{45.6} & \textbf{95.8} & \textbf{61.5} & \textbf{93.0} & \textbf{92.3} & \textbf{83.7} & \textbf{88.0} & \textbf{67.9} & \textbf{93.6} & \textbf{57.7} & \textbf{87.4} & 57.1 & \textbf{79.6} \\
    \hline
    \bottomrule
  \end{tabular}
  }
  \end{scriptsize}
  \label{tab:voc_val_detail}
\end{table}

\begin{table}[!b]
  \centering
  \caption{
   Per-class performance comparison with WSSS method in terms of IoUs (\%) on the PASCAL VOC 2012 test set.
  }
  \begin{scriptsize}
  \resizebox{\textwidth}{!}{
  \begin{tabular}{
    p{0.25\textwidth} 
    p{0.043\textwidth} p{0.043\textwidth} p{0.043\textwidth} p{0.043\textwidth} p{0.043\textwidth} 
    p{0.043\textwidth} p{0.043\textwidth} p{0.043\textwidth} p{0.043\textwidth} p{0.043\textwidth} 
    p{0.043\textwidth} p{0.043\textwidth} p{0.043\textwidth} p{0.043\textwidth} p{0.043\textwidth} 
    p{0.043\textwidth} p{0.043\textwidth} p{0.043\textwidth} p{0.043\textwidth} p{0.043\textwidth} p{0.043\textwidth} 
    c }
    \toprule
    Method & \rotatebox[origin=l]{90}{bkg} & \rotatebox[origin=l]{90}{aero} & \rotatebox[origin=l]{90}{bike} & \rotatebox[origin=l]{90}{bird} & \rotatebox[origin=l]{90}{boat} & \rotatebox[origin=l]{90}{bottle} & \rotatebox[origin=l]{90}{bus} & \rotatebox[origin=l]{90}{car} & \rotatebox[origin=l]{90}{cat} & \rotatebox[origin=l]{90}{chair} & \rotatebox[origin=l]{90}{cow} & \rotatebox[origin=l]{90}{table} & \rotatebox[origin=l]{90}{dog} & \rotatebox[origin=l]{90}{horse} & \rotatebox[origin=l]{90}{mbk} & \rotatebox[origin=l]{90}{person} & \rotatebox[origin=l]{90}{plant} & \rotatebox[origin=l]{90}{sheep} & \rotatebox[origin=l]{90}{sofa} & \rotatebox[origin=l]{90}{train} & \rotatebox[origin=l]{90}{tv}  & mIoU  \\
    \hline \hline
    EM {\tiny ICCV'15} \cite{papandreou2015weakly} & 76.3 & 37.1 & 21.9 & 41.6 & 26.1 & 38.5 & 50.8 & 44.9 & 48.9 & 16.7 & 40.8 & 29.4 & 47.1 & 45.8 & 54.8 & 28.2 & 30.0 & 44.0 & 29.2 & 34.3 & 46.0 & 39.6 \\
    MIL-LSE {\tiny CVPR'15} \cite{pinheiro2015image} & 78.7 & 48.0 & 21.2 & 31.1 & 28.4 & 35.1 & 51.4 & 55.5 & 52.8 & 7.8 & 56.2 & 19.9 & 53.8 & 50.3 & 40.0 & 38.6 & 27.8 & 51.8 & 24.7 & 33.3 & 46.3 & 40.6 \\
    SEC {\tiny ECCV'16} \cite{kolesnikov2016seed} & 83.5 & 56.4 & 28.5 & 64.1 & 23.6 & 46.5 & 70.6 & 58.5 & 71.3 & 23.2 & 54.0 & 28.0 & 68.1 & 62.1 & 70.0 & 55.0 & 38.4 & 58.0 & 39.9 & 38.4 & 48.3 & 51.7 \\
    TransferNet {\tiny CVPR'16} \cite{hong2016learning} & 85.7 & 70.1 & 27.8 & 73.7 & 37.3 & 44.8 & 71.4 & 53.8 & 73.0 & 6.7 & 62.9 & 12.4 & 68.4 & 73.7 & 65.9 & 27.9 & 23.5 & 72.3 & 38.9 & 45.9 & 39.2 & 51.2 \\
    CRF-RNN {\tiny CVPR'17} \cite{roy2017combining} & 85.7 & 58.8 & 30.5 & 67.6 & 24.7 & 44.7 & 74.8 & 61.8 & 73.7 & 22.9 & 57.4 & 27.5 & 71.3 & 64.8 & 72.4 & 57.3 & 37.3 & 60.4 & 42.8 & 42.2 & 50.6 & 53.7 \\
    WebCrawl {\tiny CVPR'17} \cite{hong2017weakly} & 87.2 & 63.9 & 32.8 & 72.4 & 26.7 & 64.0 & 72.1 & 70.5 & 77.8 & 23.9 & 63.6 & 32.1 & 77.2 & 75.3 & 76.2 & 71.5 & 45.0 & 68.8 & 35.5 & 46.2 & 49.3 & 58.7 \\
    PSA {\tiny CVPR'18} \cite{ahn2018learning} & 89.1 & 70.6 & 31.6 & 77.2 & 42.2 & 68.9 & 79.1 & 66.5 & 74.9 & 29.6 & 68.7 & 56.1 & 82.1 & 64.8 & 78.6 & 73.5 & 50.8 & 70.7 & 47.7 & 63.9 & 51.1 & 63.7 \\
    FickleNet {\tiny CVPR'19} \cite{lee2019ficklenet} & 90.3 & 77.0 & 35.2 & 76.0 & 54.2 & 64.3 & 76.6 & 76.1 & 80.2 & 25.7 & 68.6 & 50.2 & 74.6 & 71.8 & 78.3 & 69.5 & 53.8 & 76.5 & 41.8 & 70.0 & 54.2 & 65.0 \\
    SSDD {\tiny ICCV'19} \cite{shimoda2019self} & 89.5 & 71.8 & 31.4 & 79.3 & 47.3 & 64.2 & 79.9 & 74.6 & 84.9 & 30.8 & 73.5 & 58.2 & 82.7 & 73.4 & 76.4 & 69.9 & 37.4 & 80.5 & 54.5 & 65.7 & 50.3 & 65.5 \\
    RRM {\tiny AAAI'20} \cite{zhang2020reliability} & 87.8 & 77.5 & 30.8 & 71.7 & 36.0 & 64.2 & 75.3 & 70.4 & 81.7 & 29.3 & 70.4 & 52.0 & 78.6 & 73.8 & 74.4 & 72.1 & 54.2 & 75.2 & 50.6 & 42.0 & 52.5 & 62.9 \\
    SSSS {\tiny CVPR'20} \cite{araslanov2020single} & 88.7 & 70.4 & 35.1 & 75.7 & 51.9 & 65.8 & 71.9 & 64.2 & 81.1 & 30.8 & 73.3 & 28.1 & 81.6 & 69.1 & 62.6 & 74.8 & 48.6 & 71.0 & 40.1 & 68.5 & \textbf{64.3} & 62.7 \\
    SEAM {\tiny CVPR'20} \cite{wang2020self} & 88.8 & 68.5 & 33.3 & 85.7 & 40.4 & 67.3 & 78.9 & 76.3 & 81.9 & 29.1 & 75.5 & 48.1 & 79.9 & 73.8 & 71.4 & 75.2 & 48.9 & 79.8 & 40.9 & 58.2 & 53.0 & 64.5 \\
    AdvCAM {\tiny CVPR'21} \cite{lee2021anti} & 90.1 & 81.2 & 33.6 & 80.4 & 52.4 & 66.6 & 87.1 & 80.5 & 87.2 & 28.9 & 80.1 & 38.5 & 84.0 & 83.0 & 79.5 & 71.9 & 47.5 & 80.8 & 59.1 & 65.4 & 49.7 & 68.0 \\
    CPN {\tiny ICCV'21} \cite{zhang2021complementary} & 90.4 & 79.8 & 32.9 & 85.7 & 52.8 & 66.3 & 87.2 & 81.3 & 87.6 & 28.2 & 79.7 & 50.1 & 82.9 & 80.4 & 78.8 & 70.6 & 51.1 & 83.4 & 55.4 & 68.5 & 44.6 & 68.5 \\
    RIB {\tiny NeurIPS'21} \cite{lee2021reducing} & 90.4 & 80.5 & 32.8 & 84.9 & 59.4 & 69.3 & 87.2 & 83.5 & 88.3 & 31.1 & 80.4 & 44.0 & 84.4 & 82.3 & 80.9 & 70.7 & 43.5 & 84.9 & 55.9 & 59.0 & 47.3 & 68.6 \\
    AMN {\tiny CVPR'22} \cite{lee2022threshold} & 90.7 & 82.8 & 32.4 & 84.8 & 59.4 & 70.0 & 86.7 & 83.0 & 86.9 & 30.1 & 79.2 & 56.6 & 83.0 & 81.9 & 78.3 & 72.7 & 52.9 & 81.4 & 59.8 & 53.1 & 56.4 & 69.6 \\
    W-OoD {\tiny CVPR'22} \cite{lee2022weakly} & 91.4 & 85.3 & 32.8 & 79.8 & 59.0 & 68.4 & 88.1 & 82.2 & 88.3 & 27.4 & 76.7 & 38.7 & 84.3 & 81.1 & 80.3 & 72.8 & 57.8 & 82.4 & 59.5 & {79.5} & 52.6 & 69.9 \\
    RCA {\tiny CVPR'22} \cite{zhou2022regional} & 92.1 & 86.6 & 40.0 & 90.1 & 60.4 & 68.2 & 89.8 & 82.3 & 87.0 & 27.2 & 86.4 & 32.0 & 85.3 & 88.1 & 83.2 & 78.0 & 59.2 & 86.7 & 45.0 & 71.3 & 52.5 & 71.0 \\
    SANCE {\tiny CVPR'22} \cite{li2022towards} & 91.6 & 82.6 & 33.6 & 89.1 & 60.6 & \textbf{76.0} & 91.8 & 83.0 & 90.9 & 33.5 & 80.2 & 64.7 & 87.1 & 82.3 & 81.7 & 78.3 & 58.5 & 82.9 & {60.9} & 53.9 & 53.5 & 72.2 \\
    MCTformer {\tiny CVPR'22} \cite{xu2022multi} & 92.3 & 84.4 & 37.2 & 82.8 & 60.0 & 72.8 & 78.0 & 79.0 & 89.4 & 31.7 & 84.5 & 59.1 & 85.3 & 83.8 & 79.2 & 81.0 & 53.9 & 85.3 & 60.5 & 65.7 & 57.7 & 71.6 \\
    RSEPM {\tiny arXiv'22} \cite{jo2022recurseed} & 91.9 & 89.7 & 37.3 & 88.0 & 62.5 & 72.1 & 93.5 & 85.6 & 90.2 & 36.3 & {88.3} & 62.5 & 86.3 & {89.1} & 82.9 & 81.2 & 59.7 & 89.2 & 56.2 & 44.5 & 59.4 & 73.6 \\
    MARS {\tiny ICCV'23} & {93.7} & \textbf{93.3} & {40.3} & \textbf{90.8} & {70.8} & 71.7 & {94.0} & {86.3} & {93.9} & {40.4} & 87.6 & {67.6} & {90.0} & 87.3 & {83.9} & {83.1} & {64.2} & {89.5} & 59.6 & 79.0 & 55.1 & {77.2} \\
    \rowcolor[HTML]{F4CCCC}
    \textbf{DHR (ResNet-101)} & \textbf{94.2} & \textbf{93.3} & \textbf{42.6} & {86.6} & \textbf{74.8} & 72.3 & \textbf{95.0} & \textbf{88.3} & \textbf{95.1} & \textbf{41.6} & \textbf{90.9} & \textbf{71.2} & \textbf{93.3} & \textbf{93.3} & \textbf{86.8} & \textbf{85.7} & \textbf{73.9} & \textbf{93.9} & \textbf{63.4} & \textbf{81.8} & 56.8 & \textbf{79.8} \\
    \hline
    \bottomrule
  \end{tabular}
  }
  \end{scriptsize}
  \label{tab:voc_test_detail}
\end{table}

\clearpage

\clearpage

\clearpage

\clearpage

%
%
\bibliographystyle{splncs04}
\bibliography{main}
\end{document}